\documentclass[11pt,twoside,twocolumn,a4paper]{article}

\usepackage{cvww}
\usepackage{times}
\usepackage{epsfig}
\usepackage{graphicx}
\usepackage{amsmath}
\usepackage{amssymb}

\usepackage{mathtools}
\usepackage{verbatim}
\usepackage{caption}
\usepackage{subcaption}
\usepackage{tabularx}
\usepackage{colortbl}
\usepackage{makecell}
\usepackage{etoolbox}
\usepackage{booktabs}
\usepackage{arydshln}
\usepackage{colortbl}
\usepackage{tabu}
\usepackage{multirow}
\usepackage{algorithm}
\usepackage[noend]{algpseudocode}
\usepackage{enumitem}
\usepackage[dvipsnames]{xcolor}
\usepackage[skip=1ex,font=small,labelsep=period]{caption}

\newcolumntype{Y}{>{\centering\arraybackslash}X}
\newcolumntype{R}{>{\raggedright\arraybackslash}X}
\newcolumntype{L}{>{\raggedleft\arraybackslash}X}

\definecolor{g1}{rgb}{1.00,1.00,1.00}
\definecolor{lightgray}{rgb}{0.9, 0.9, 0.9}
\definecolor{wincolor}{rgb}{0.95, 0.2, 0.2}

\newcommand{\win}[1]{\textcolor{wincolor}{\textbf{#1}}}
\newcommand{\second}[1]{\textcolor{NavyBlue}{\textbf{#1}}}
\newcommand{\dataset}[1]{{\fontfamily{cmtt}\selectfont #1}}

\makeatletter
\def\adl@drawiv#1#2#3{%
        \hskip.5\tabcolsep
        \xleaders#3{#2.5\@tempdimb #1{1}#2.5\@tempdimb}%
                #2\z@ plus1fil minus1fil\relax
        \hskip.5\tabcolsep}
\newcommand{\cdashlinelr}[1]{%
  \noalign{\vskip\aboverulesep
           \global\let\@dashdrawstore\adl@draw
           \global\let\adl@draw\adl@drawiv}
  \cdashline{#1}
  \noalign{\global\let\adl@draw\@dashdrawstore
           \vskip\belowrulesep}}
\makeatother

\makeatletter
\newlength{\qrr@dimen@}
\expandafter\pretocmd\csname tabular*\endcsname{\setlength{\qrr@dimen@}{#1}}{}{}
\newcommand*{\Rowcolor}[2][\tabcolsep]{%
    \ifx\relax#1\relax\else
        \kern-\the\dimexpr#1\relax
    \fi
    \makebox[0pt][l]{%
        \fboxsep=0pt
        \colorbox{#2}{%
            \strut\kern\qrr@dimen@
        }%
    }%
    \ifx\relax#1\relax\else
        \kern\the\dimexpr#1\relax
    \fi
    \ignorespaces
}
\makeatother
           
\newcommand\customparagraph[1]{\vspace{0.6em}\noindent\textbf{#1}}

\usepackage[pagebackref=true,breaklinks=true,letterpaper=true,colorlinks,bookmarks=false]{hyperref}

\usepackage[pagebackref=true,breaklinks=true,bookmarks=false]{hyperref}
\cvwwfinalcopy 

\def\cvwwPaperID{24} 

\begin{document}

\title{ USACv20: robust essential, fundamental and homography matrix estimation }

\author{Maksym Ivashechkin$^{1}$, Daniel Barath$^{12}$, and Jiri Matas$^{1}$\\
$^1$ Centre for Machine Perception, Czech Technical University in Prague, Czech Republic\\
$^2$ Machine Perception Research Laboratory, MTA SZTAKI, Budapest, Hungary\\
{\tt\small \{ivashmak, matas\}@cmp.felk.cvut.cz} \:\:\: {\tt\small barath.daniel@sztaki.mta.hu}\\
}


\maketitle
\ifcvwwfinal\thispagestyle{fancy}\fi


\begin{abstract}
    We review the most recent RANSAC-like hypothesize-and-verify robust estimators. 
    The best performing ones are combined to create a state-of-the-art version of the Universal Sample Consensus (USAC) algorithm.
    A recent objective is to implement a modular and optimized framework, making future RANSAC modules easy to be included.
    The proposed method, USACv20, is tested on eight publicly available real-world datasets, estimating homographies, fundamental and essential matrices. 
    On average, USACv20 leads to the most geometrically accurate models and it is the fastest in comparison to the state-of-the-art robust estimators.
    All reported properties improved performance of original USAC algorithm significantly.
    %
    The pipeline will be made available after publication.
\end{abstract}


\begin{figure}[t]
    \centering
   \begin{subfigure}[t]{0.99\columnwidth}
\includegraphics[width=1.0\columnwidth]{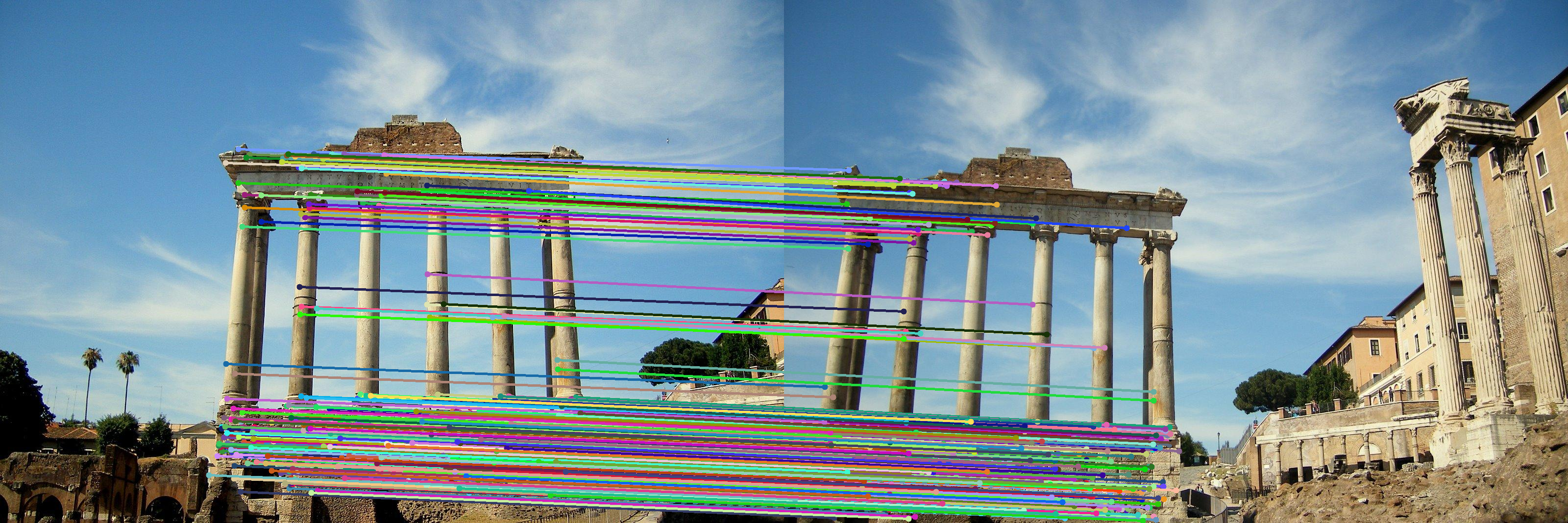}
        \caption{\dataset{Community Photo Collection dataset}~\cite{wilson2014robust}. }
   \end{subfigure}
   \begin{subfigure}[t]{0.99\columnwidth}
        \includegraphics[width=1.0\columnwidth]{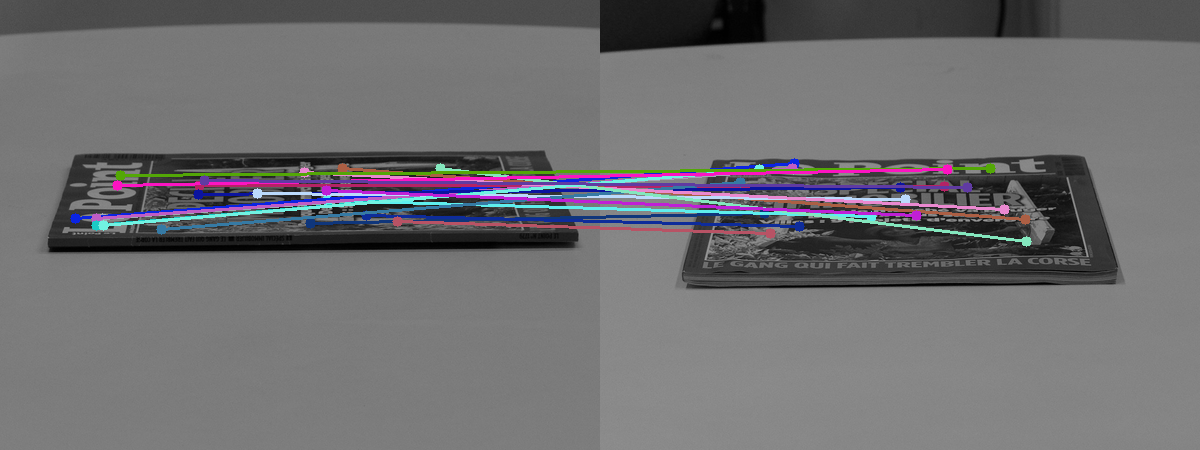}
        \caption{\dataset{ExtremeView dataset}~\cite{mishkin2015mods}. }
   \end{subfigure}
   \begin{subfigure}[t]{0.99\columnwidth}
        \includegraphics[width=1.0\columnwidth]{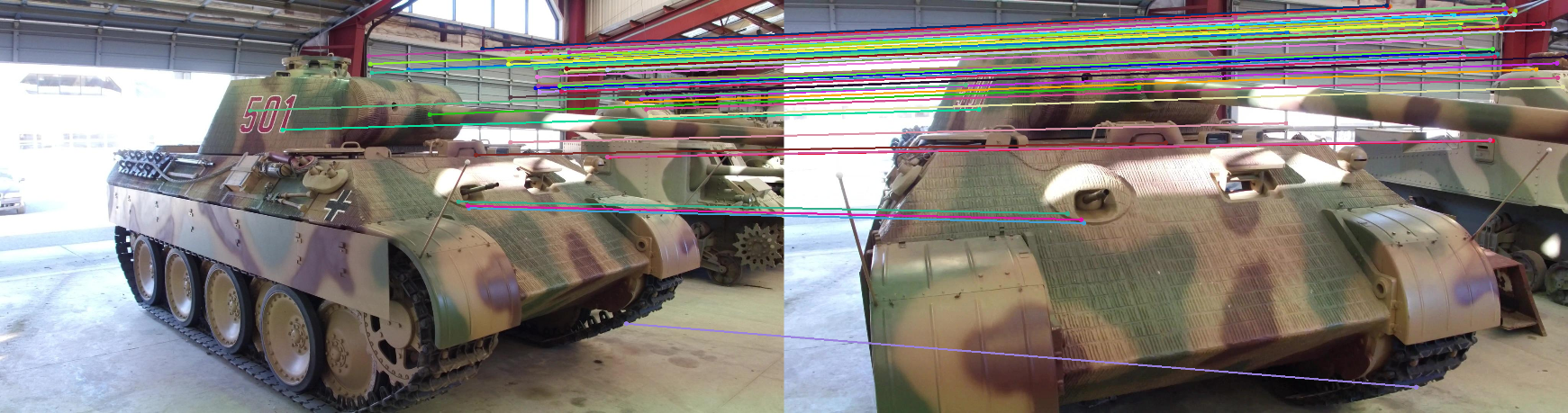}
        \caption{\dataset{Tanks and Temples dataset}~\cite{knapitsch2017tanks}.}
   \end{subfigure}
   
   \begin{subfigure}[t]{0.99\columnwidth}
    \includegraphics[width=1.0\columnwidth]{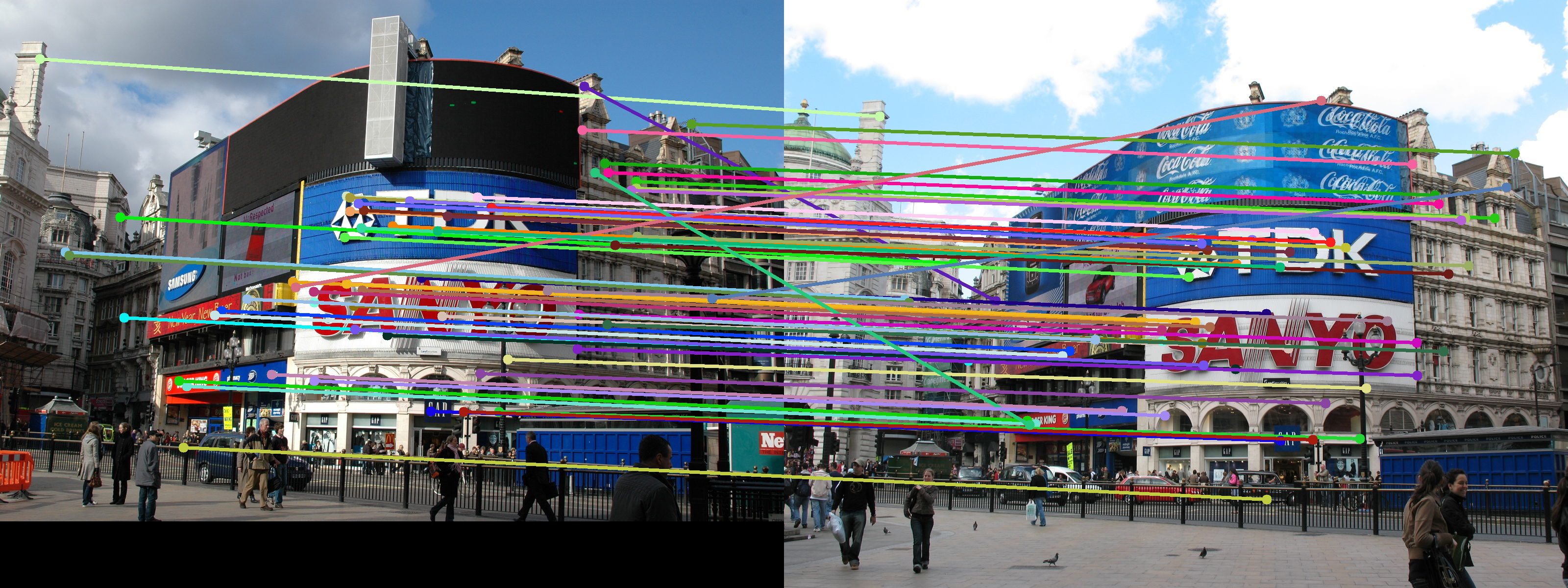}
        \caption{\dataset{Piccadilly dataset}~\cite{wilson2014robust}. }
   \end{subfigure}
   
    \caption{Example image pairs where USACv20 has lower error to ground truth inliers than OpenCV RANSAC and USAC \cite{raguram2013usac} estimators.}
    \label{fig:example_result}
\end{figure}

\section{Introduction}

The RANdom SAmple Consensus (RANSAC) algorithm~\cite{fischler1981random} has been one of the most widely used robust estimators in computer vision. 
RANSAC and many of its variants have been successfully applied to a wide range of vision tasks, for instance, short baseline stereo~\cite{torr1993outlier,torr1998robust}, motion segmentation~\cite{torr1993outlier}, detection of geometric primitives~\cite{sminchisescu2005incremental}, wide baseline matching~\cite{pritchett1998wide,matas2004robust,mishkin2015mods}, in structure-from-motion~\cite{agarwal2010bundle,wilson2014robust, schonberger2016structure} (SfM) or simultaneous localization and mapping~\cite{engel2014lsd,mur2015orb} (SLAM) pipelines, image mosaicing~\cite{ghosh2016survey}, and to perform~\cite{zuliani2005multiransac} or initialize multi-model fitting~\cite{isack2012energy,pham2014interacting}.

In this paper, we review some of the most recent RANSAC modifications, combine them together and propose a state-of-the-art variant of the Universal Sample Consensus~\cite{raguram2013usac} (USAC) algorithm. 
Also, an important objective is to make the implemented modular and optimized C++ framework publicly available, therefore, making future RANSAC modules easy to be combined with the proposed USACv20.

In short, the RANSAC approach repeatedly creates minimal sets of randomly selected points and fits a model to them, e.g., a circle to three 2D points or a homography to four 2D point correspondences.
Next, the quality of the estimated model is measured, for example, by the cardinality of its support, i.e., the number of data points closer than a manually set inlier-outlier threshold. 
Finally, the model with the highest score, polished, e.g., by least squares fitting of all inliers, is returned.

\textit{Scoring function.} Many modifications have been proposed since the publication of RANSAC, improving the components of the algorithm.
For instance, in MAPSAC~\cite{torr2002bayesian}, the robust estimation is formulated as a process that estimates both the parameters of the data distribution and the quality of the model in terms of maximum a posteriori. 
MLESAC~\cite{torr2000mlesac} estimates the model quality by a maximum likelihood process with all its beneficial properties, albeit under certain assumptions about data distributions. 
In practice, MLESAC results are often superior to the inlier counting of plain RANSAC, and are less sensitive to the inlier-outlier threshold defined manually.

\textit{Local Optimization.} Observing that RANSAC requires in practice  more samples than theory predicts, Chum et al.~\cite{chum2003locally,lebeda2012fixing} identified a problem that not all all-inlier samples are ``good'', i.e., lead to a model accurate enough to distinguish all inliers, e.g., due to poor conditioning of the selected random all-inlier sample. They addressed the problem by introducing the locally optimized RANSAC that augments the original approach with a local optimization step applied to the \textit{so-far-the-best} model. 
This approach had been further improved in Graph-Cut RANSAC~\cite{barath2018graph} considering the fact that real-world data often form spatially coherent structures.
Graph-Cut RANSAC exploits the proximity of the points in the local optimization step, leading to results superior to LO-RANSAC in terms of geometric accuracy.

\textit{Sampling Strategies.}
Samplers NAPSAC~\cite{nasuto2002napsac} and PROSAC~\cite{chum2005matching} modify the RANSAC sampling strategy to increase the probability of selecting an all-inlier sample early. 
PROSAC exploits an a priori predicted inlier probability rank of the points and starts the sampling with the most promising ones.
PROSAC and other RANSAC-like samplers treat models without considering
that inlier points often are in the proximity of each other. 
This approach is effective when finding a global model with inliers sparsely distributed in the scene, for instance, the rigid motion induced by changing the viewpoint in two-view matching.
However, as it is often the case in real-world data, if the model is localized with inlier points close to each other, robust estimation can be significantly sped up by exploiting this in the sampling.
NAPSAC assumes that inliers are spatially coherent.
It draws samples from a hyper-sphere centered at the first, randomly selected, point. 
If this point is an inlier, the rest of the points sampled in its proximity are more likely to be inliers than the points outside the ball.
Progressive NAPSAC~\cite{barath2019progressive} was proposed to combine NAPSAC-like localized sampling with PROSAC by drawing minimal samples from gradually growing neighborhoods. 

\textit{Optimizing Model Verification.}
One of the most successful improvement for speeding up the verification is the optimal randomized model verification strategy~\cite{matas2005randomized,chum2008optimal} (WaldSAC) based on Wald’s theory of sequential decision making.
When the level of outlier contamination is known a priori, the WaldSAC strategy is provably optimal. In practice, however, inlier ratios have to be estimated during the evaluation process and WaldSAC adjusted to the current so-far-the-best model.
The performance of the SPRT test is not significantly affected by the imperfect estimation of these parameters. 

\textit{Termination criterion.}
There were a number of different termination criteria proposed for RANSAC-like hypothesize-and-verify methods. 
The original criterion is based on the assumption that the inliers are noise-free. The number of iterations required is calculated from the inlier ratio and the number of points needed for the model estimation. 
This criterion was then relaxed by Progressive NAPSAC~\cite{barath2019progressive} by terminating if the probability of finding a model which has significantly more inliers than the previous best falls below a threshold. 
In \cite{chum2005matching}, another criterion was proposed. 
The PROSAC algorithm terminates if the number of inliers satisfies the following conditions:
(i) non-randomness -- the probability that $i^*$ out of $n$ data points are by chance inliers to an arbitrary incorrect model is smaller than a threshold; (ii) maximality -- the probability that a solution with more than $i^*$ inliers exists and was not found after $k$ samples is smaller than $\mu_0$.

\begin{algorithm}
\begin{algorithmic}[1]
    \Statex{\hspace{-1.1em}\textbf{Input:} $\mathcal{P}$ -- points; $\eta$ -- confidence, $t$ -- maximum iterations, $\mathcal{T}$ -- termination, ...}
   \Statex{\hspace{-1.1em}\textbf{Output:} $\hat{\theta}^{*}$ -- the best found model}
    \State{$\varepsilon^* \leftarrow \infty$} 
    \While{! \emph{terminate ($\mathcal{T}, \eta, t$)}} \label{usac:line:term}
        \State{$\mathcal{S} \leftarrow $ \emph{sampling} $(\mathcal{P}$)} \label{usac:line:sampl}

        \If{! \emph{validate\_sample} ($\mathcal{S}$)}
            \State{\bf \emph{continue}} 
        \EndIf
        \State{$\hat{\Theta} \leftarrow$ \emph{estimate} $(\mathcal{S})$}
        
        \For{$\hat{\theta} \in \hat{\Theta}$}
            \If {! \emph{validate\_model ($\hat{\theta}, \mathcal{S}$)}} \label{usac:line:model_degen}
                \State{\bf \emph{continue}}
             \EndIf
             \If{! \emph{preemptive\_verification($\hat{\theta}$)}} \label{usac:line:preemp_ver}
                 \State{\bf \emph{continue}}
           \EndIf
        
            \State{$\varepsilon \leftarrow$ \emph{model\_quality}$(\hat{\theta})$} \label{usac:line:quality}
             
             \If {$\varepsilon^{*} \prec \varepsilon$} 
                \State{$\hat{\theta}^{'} \leftarrow$ \emph{recover\_if\_degenerate} $(\hat{\theta}, \mathcal{S})$} \label{usac:line:model_degen_F}

                \If {$\hat{\theta}^{'} = $ NULL}
                    \State{\bf \emph{continue}}
                \EndIf

                \State{$\varepsilon^{'} \leftarrow $ \emph{model\_quality} ($\hat{\theta^{'}}$)}
                
                \If {$\varepsilon^{*} \prec \varepsilon^{'}$}
                    \State{$\hat{\theta}_{LO} \leftarrow$ \emph{local\_optimization} ($\hat{\theta}^{'}$)} \label{usac:line:lo}
                    \State{$\hat{\theta}_{LO} \leftarrow$ \emph{recover} ($\hat{\theta}_{LO}$)}

                    \If {$\hat{\theta}_{LO} \ne $ NULL}
                    \State{$\varepsilon_{LO} \leftarrow$ \emph{model\_quality} ($\hat{\theta}_{LO}$)}
                    
                     \If {$\varepsilon^{'} \prec \varepsilon_{LO}$}
                            \State{$\hat{\theta}^{'}, \varepsilon^{'} \leftarrow \hat{\theta}_{LO}, \varepsilon_{LO}$}
                    \EndIf    
                    
                     \EndIf

                     \State{$\hat{\theta}^{*}, \varepsilon^{*} \leftarrow \hat{\theta}^{'}, \varepsilon^{'}$}
                    
                    \State{$\mathcal{T} \leftarrow $ \emph{update} $(\hat{\theta}^*, \mathcal{I}_{\hat{\theta}^*})$}
                \EndIf
                
            \EndIf
        \EndFor
    \EndWhile
    \State{$\hat{\theta}^* \leftarrow$ \emph{polish\_final} $(\hat{\theta}^*)$} \label{usac:line:final_polishing}

\end{algorithmic}
\caption{\bf USACv20. }
\label{alg:usacv20}
\end{algorithm}

\section{USACv20}
The structure of the proposed framework is summarized in Algorithm \ref{alg:usacv20}. The standard RANSAC loop is executed between lines \ref{usac:line:term}: and \ref{usac:line:final_polishing}:. The implementation is modular, and each step of the algorithm allows a range of options.

In the version of USACv20 evaluated in the paper, the chosen sampling method is Progressive NAPSAC, alg. \ref{alg:usacv20}, line \ref{usac:line:sampl}. Other samplers are described in section \ref{sampling}. The pre-emptive model verification is SPRT, alg. \ref{alg:usacv20}, line \ref{usac:line:preemp_ver}. Other options could be none verification or $T_{d,d}$ test, see section \ref{verification}. The termination condition, alg. \ref{alg:usacv20}, line \ref{usac:line:term} is combination of SPRT and P-NAPSAC since P-NAPSAC and SPRT are used. The measured quality of model is MSAC (sum of truncated errors), alg. \ref{alg:usacv20}, line \ref{usac:line:quality}. The MSAC quality could be also replaced by MLESAC or MAGSAC quality, see section \ref{quality}. The local optimization step is done in the line \ref{usac:line:lo} by graph-cut-based local optimization. Other modifications of local optimization are in the section \ref{local_optimization}.

The degeneracy of model (e.g., validation of epipolar oriented constraint \cite{chum2004epipolar}) is done in the alg. \ref{alg:usacv20}, line \ref{usac:line:model_degen} and after finding so-far-the-best model in the line \ref{usac:line:model_degen_F} (e.g., planarity of fundamental matrix \cite{chum2005two}. In the end the output model is polished by least squares fitting on all inliers, alg. \ref{alg:usacv20}, line \ref{usac:line:final_polishing}.

\subsection{Local optimization} \label{local_optimization}

The options for local optimization are listed below.
The one chosen in USACv20 is written in bold.
\begin{center}
\begin{tabular}{ | m{2.5cm} | m{4.5cm} | } 
\hline
LO-RANSAC \cite{chum2003locally} & Refine each so-far-the-best model by an inner RANSAC. \\  \hline
FLO-RANSAC \cite{lebeda2012fixing} & Improvement of LO-RANSAC. \\  \hline
\bf{Graph-Cut RANSAC} \cite{barath2018graph} & Spatial coherence is considered when doing the inner RANSAC. \\ \hline
$\sigma$-consensus \cite{barath2019magsac} & A part of the MAGSAC algorithm marginalizing over the noise-scale. \\  \hline
\end{tabular}
\end{center}
We chose Graph-Cut RANSAC since it is more accurate than LO-RANSAC and FLO-RANSAC and significantly faster than the $\sigma$-consensus which requires a number of least-squares fittings.




\subsection{Sampling} \label{sampling}

The possible options for sampling are listed below.
The one chosen in USACv20 is written in bold.
\begin{center}
\begin{tabular}{ | m{2.5cm} | m{4.5cm} | } 
\hline
Uniform \cite{fischler1981random} & The default option. \\  \hline
NAPSAC \cite{nasuto2002napsac} & Selecting the first points and, then, local sampling from its neighborhood. \\ \hline
PROSAC \cite{chum2005matching} & Sampling from the most promising samples first and progressively blending to the uniform sampler of RANSAC. \\ \hline
\bf{P-NAPSAC} \cite{barath2019progressive} & Combination of PROSAC and NAPSAC sampling from gradually growing neighborhoods. \\ \hline
\end{tabular}
\end{center}
We chose P-NAPSAC since it leads to finding a good-enough sample earlier than PROSAC when the sought model is localized. In case of having a global model, e.g.\ the background motion in two images, it is found not noticeably later than by PROSAC due to progressively blending into global sampling.




\subsection{Quality} \label{quality}

The options for the model quality calculation are listed below.
The one chosen is written in bold.
\begin{center}
\begin{tabular}{ | m{2.5cm} | m{4.5cm} | } 
\hline
RANSAC \cite{fischler1981random} & The number of inliers. \\  \hline
\bf{MSAC} \cite{torr2000mlesac} & The sum of truncated errors. \\ \hline
MLESAC \cite{torr2000mlesac} & Likelihood of the model. \\ \hline
LMedS \cite{Rousseeuw84leastmedian} & The least median of errors. \\ \hline
MAGSAC \cite{barath2019magsac} & Sum of errors marginalized over the noise-scale. \\ \hline
\end{tabular}
\end{center}
We chose MSAC quality calculation since it is always more accurate than that of RANSAC; it does not require expensive calculations like MLESAC or MAGSAC; and does not need to know the outlier ratio a priori as LMedS does.

\subsection{Pre-emptive verification} \label{verification}

The options for the pre-emptive verification are listed below.
The one chosen is written in bold.
\begin{center}
\begin{tabular}{ | m{2.5cm} | m{4.5cm} | } 
\hline
$T_{d,d}$ \cite{chum2008optimal} & If $d$ out $d$ points are inliers then model is good. \\  \hline
\bf{SPRT} \cite{chum2008optimal} & Verify model by sequential decision making based on Wald's theory. \\ \hline
\end{tabular}
\end{center}
The $T_{d,d}$ test can make many false-negatives (rejecting good models) when the inlier ratio is low. Therefore we chose SPRT verification.



\subsection{Termination criterion} \label{termination}

The options for the termination criterion are listed below.
The one chosen is written in bold.
\begin{center}
\begin{tabular}{ | m{2.5cm} | m{4.5cm} | } 
\hline
Standard \cite{fischler1981random} & Terminates if the probability of finding a model with more inliers than the previous best falls below a threshold with some confidence. \\  \hline
PROSAC \cite{chum2005matching} & Terminates when the maximality and non-randomness criteria are satisfied. \\ \hline
\bf{SPRT} \cite{chum2008optimal} & Termination based on a sequence of subsequent model validations. \\ \hline
\bf{P-NAPSAC} \cite{barath2019progressive} & The standard RANSAC criterion relaxed by requiring the new model to select significantly more inliers than the previous best. \\ \hline
MAGSAC \cite{barath2019magsac} & Marginalization of the standard RANSAC criterion over the noise-scale $\sigma$.  \\ \hline
\end{tabular}
\end{center}
The termination of SPRT and P-NAPSAC depends on different properties of the robust procedure. P-NAPSAC stops when the relaxed RANSAC criterion is triggered, meaning that the probability of finding a significantly better model than the previous best falls below a threshold.
The SPRT criterion is triggered by the number of subsequent model verification sequences made.
These two techniques can straightforwardly be combined.
Thus, we stop when at least one of them is triggered. 






\subsection{Degeneracy} \label{degeneracy}
USACv20 framework includes different tests on degeneracy. 
DEGENSAC \cite{chum2005two} is about detecting when the majority of the drawn sample originates from the same 3D plane. 
For fundamental and essential matrix estimation oriented epipolar constraint \cite{chum2004epipolar} is evaluated as well. For homography estimation the verification of samples by its orientation is included.


\subsection{Other features}
For PROSAC or Progressive NAPSAC, exploiting an a priori known quality of the input data points makes the finding of a good-enough model significantly earlier than by other samplers. 
However, such prior information usually is unknown, degrading PROSAC to being the entirely uniform sampler of RANSAC. 
In the proposed USACv20 framework, when such quality function is not available, we use the density of the points as the quality function.
This reflects the fact real-world data often forms spatially coherent structures and, thus, good correspondences tend to be close. 

The spatial coherence of points plays important role in the estimation. 
For instance, it is exploited in the graph-cut-based local optimization or in P-NAPSAC sampler. Consequently, the neighborhood graph must be computed. The efficient way to do this is using a multi-layer grid described in \cite{barath2019progressive}. In USACv20 such neighborhood estimation is implemented and used in the experiments.

\section{Experimental results}
We compared the proposed USACv20 to three robust estimators, i.e., USAC \cite{raguram2013usac} \footnote{\url{http://wwwx.cs.unc.edu/~rraguram/usac/USAC-1.0.zip}}, GC-RANSAC \cite{barath2018graph} and the RANSAC implementation of OpenCV. 
The applied USACv20 consists of SPRT verification, DEGENSAC \cite{chum2005two}, P-NAPSAC sampler and the local optimization of GC-RANSAC. 
USAC estimator \cite{raguram2013usac} includes SPRT verification, DEGENSAC, PROSAC sampler and the local optimization of the original LO-RANSAC. 
All estimators were tested using the same number of maximum iterations (10,000 for $\textbf{H}$ and 1,000 for $\textbf{F}, \textbf{E}$ estimation) and confidence equals to 99\%.

\customparagraph{Fundamental matrix} estimation was evaluated on the benchmark of~\cite{bian2019evaluation}.
The \cite{bian2019evaluation} benchmark includes: (1) the \dataset{TUM} dataset~\cite{sturm2012benchmark} consisting of videos of indoor scenes. Each video is of resolution $640 \times 480$.
(2) The \dataset{KITTI} dataset~\cite{geiger2012we} consists of consecutive frames of a camera mounted to a moving vehicle. The images are of resolution $1226 \times 370$.
Both in \dataset{KITTI} and \dataset{TUM}, the image pairs are short-baseline.
(3) The \dataset{Tanks and Temples} (\dataset{T\&T}) dataset~\cite{knapitsch2017tanks} provides images of real-world objects for image-based reconstruction and, thus, contains mostly wide-baseline pairs. 
The images are of size from $1080 \times 1920$ up to $1080 \times 2048$.
(4) The \dataset{Community Photo Collection} (\dataset{CPC}) dataset~\cite{wilson2014robust} contains images of various sizes of landmarks collected from Flickr. 
In the benchmark, $1\,000$ image pairs are selected randomly from each dataset. SIFT~\cite{lowe1999object}
correspondences are detected, filtered by the standard SNN ratio test~\cite{lowe1999object} and, finally, used for estimating the epipolar geometry.

The compared methods are USAC~\cite{raguram2013usac}, GC-RANSAC~\cite{barath2018graph}, the RANSAC~\cite{fischler1981random} implementation in OpenCV and the proposed USACv20.
For all methods, the confidence was set to $0.99$. 
For each method and problem, we chose the threshold maximizing the accuracy.
The used error metric is Sampson distance.
All methods were in C++.

The first four blocks of Table~\ref{tab:magsac_real_experiments} report the median errors ($\epsilon_{\text{med}}$, in pixels), the failure rates ($f$; in percentage) and processing times ($t$; in milliseconds) on the datasets used for fundamental matrix estimation.
We report the median values to avoid being affected by the failures -- which are also shown. 
A test is considered failure if the error of the estimated model is bigger than the $1\%$ of the image diagonal.
The best values are shown in red, the second best ones are in blue. 
It can be seen that \textit{USACv20 leads to the lowest errors} on all datasets. Its failure ratio and processing time are always the lowest or the second lowest. 

\begin{figure}[t]
  \centering
  \includegraphics[width=0.49\columnwidth]{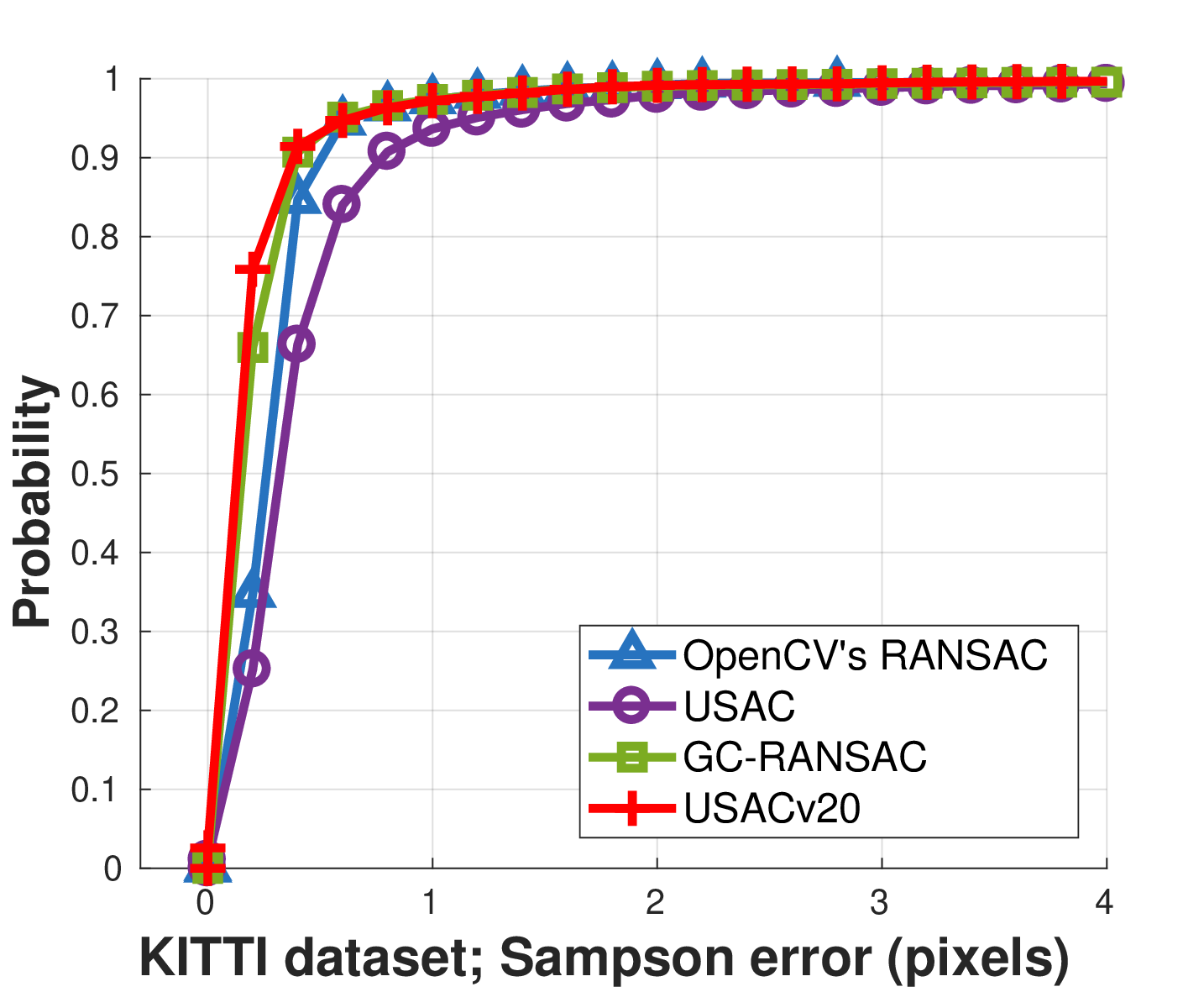}
  \includegraphics[width=0.49\columnwidth]{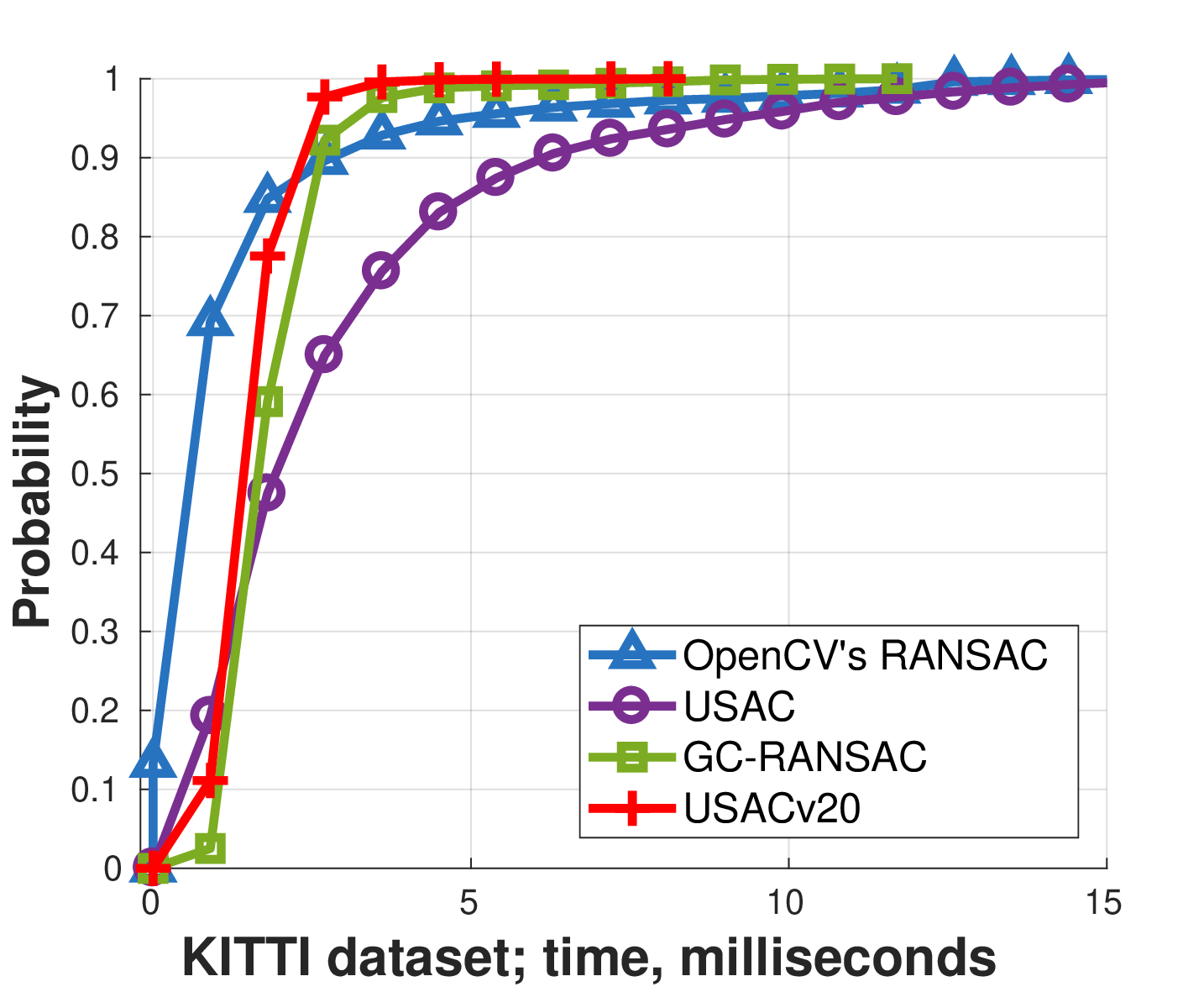}
  \caption{The cumulative distribution functions (CDF) of the Sampson errors (left plot; horizontal axis) and processing times (right; milliseconds) of the estimated fundamental matrices on the \dataset{KITTI} dataset.}
\label{fig:F_kitti}
\end{figure}

\begin{figure}[t]
  \centering
  \includegraphics[width=0.49\columnwidth]{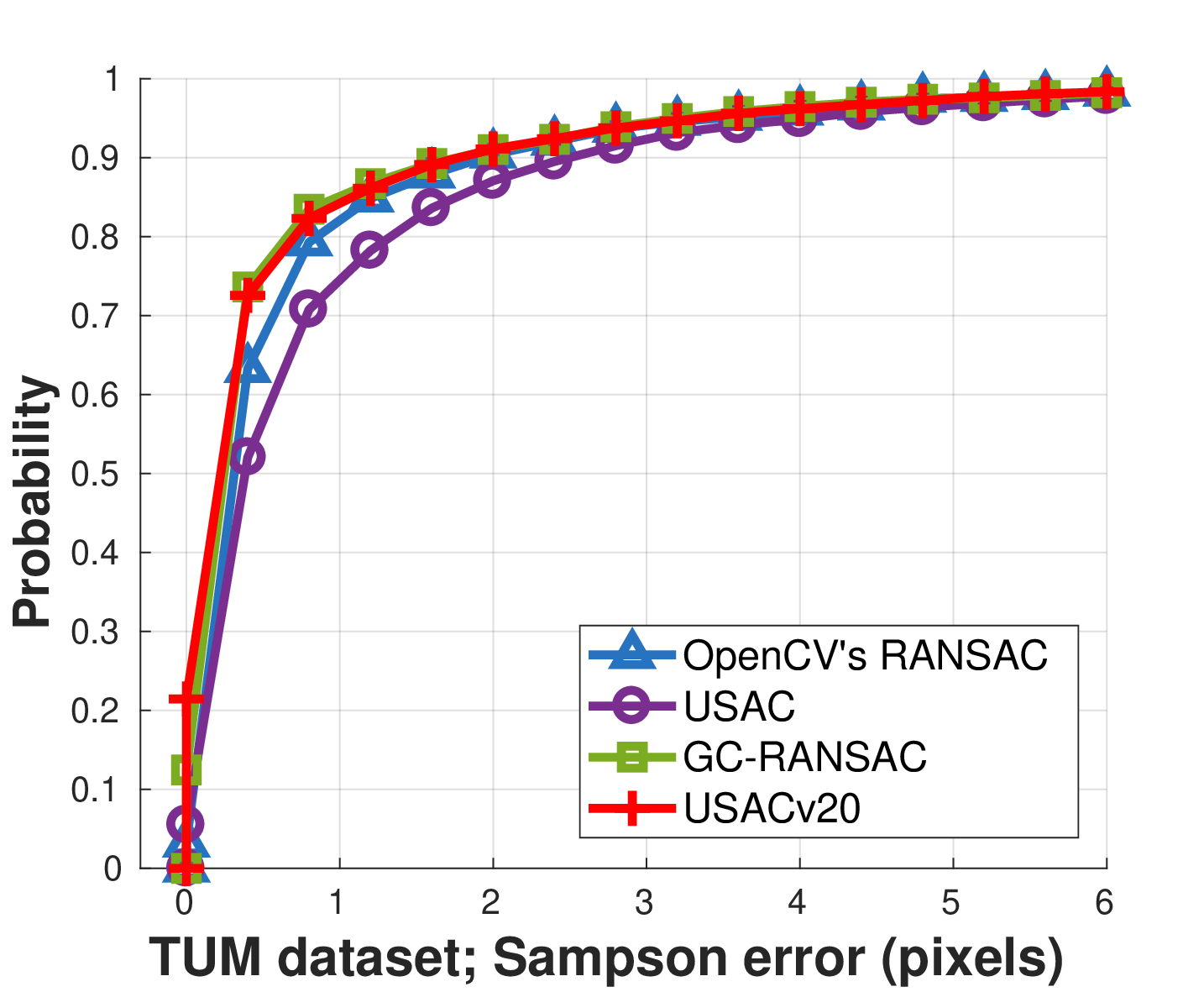}
  \includegraphics[width=0.49\columnwidth]{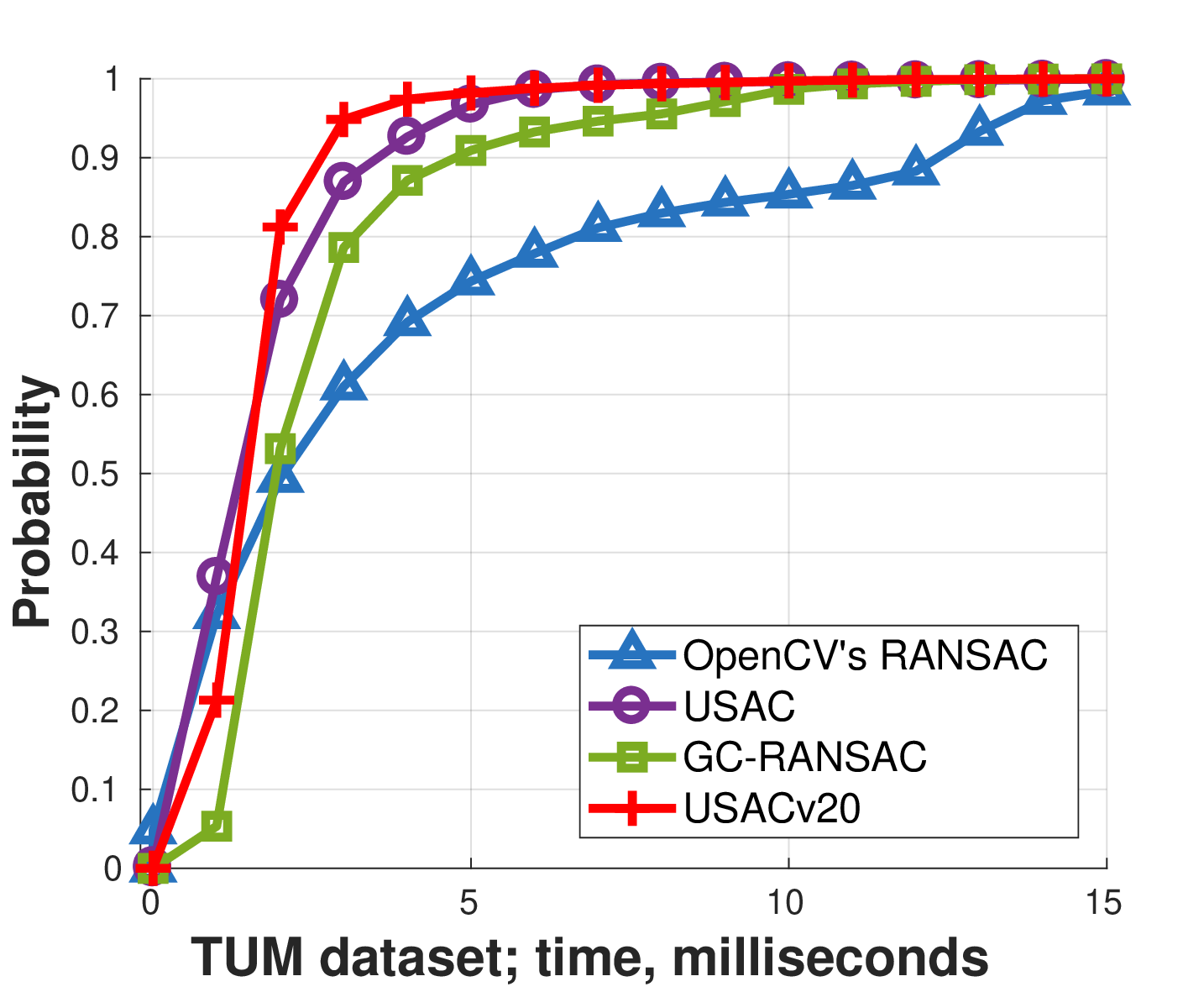}
  \caption{The cumulative distribution functions (CDF) of the Sampson errors (left plot; horizontal axis) and processing times (right; milliseconds) of the estimated fundamental matrices on the \dataset{TUM} dataset.}
\label{fig:F_tum}
\end{figure}

\begin{figure}[t]
  \centering
  \includegraphics[width=0.49\columnwidth]{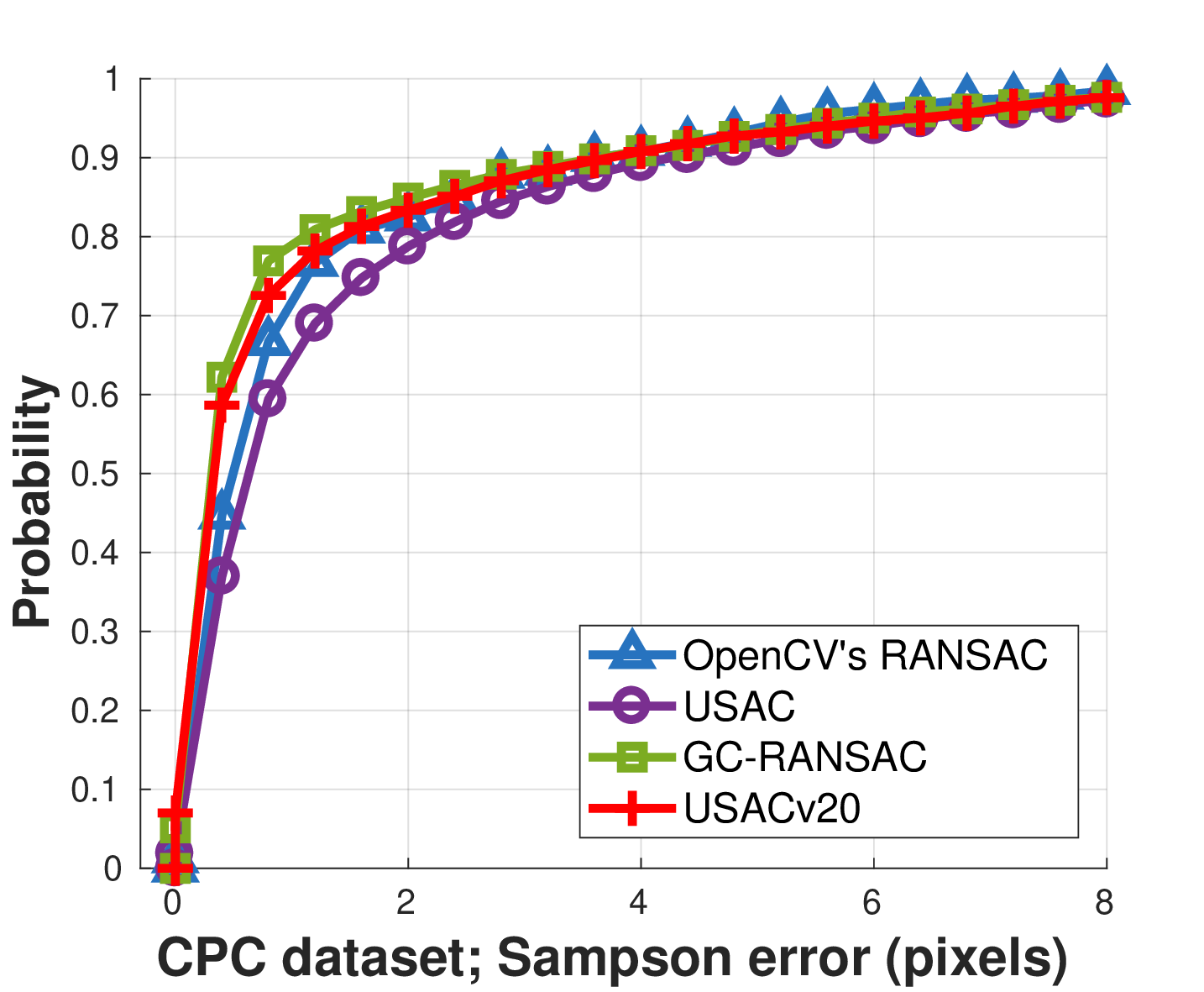}
  \includegraphics[width=0.49\columnwidth]{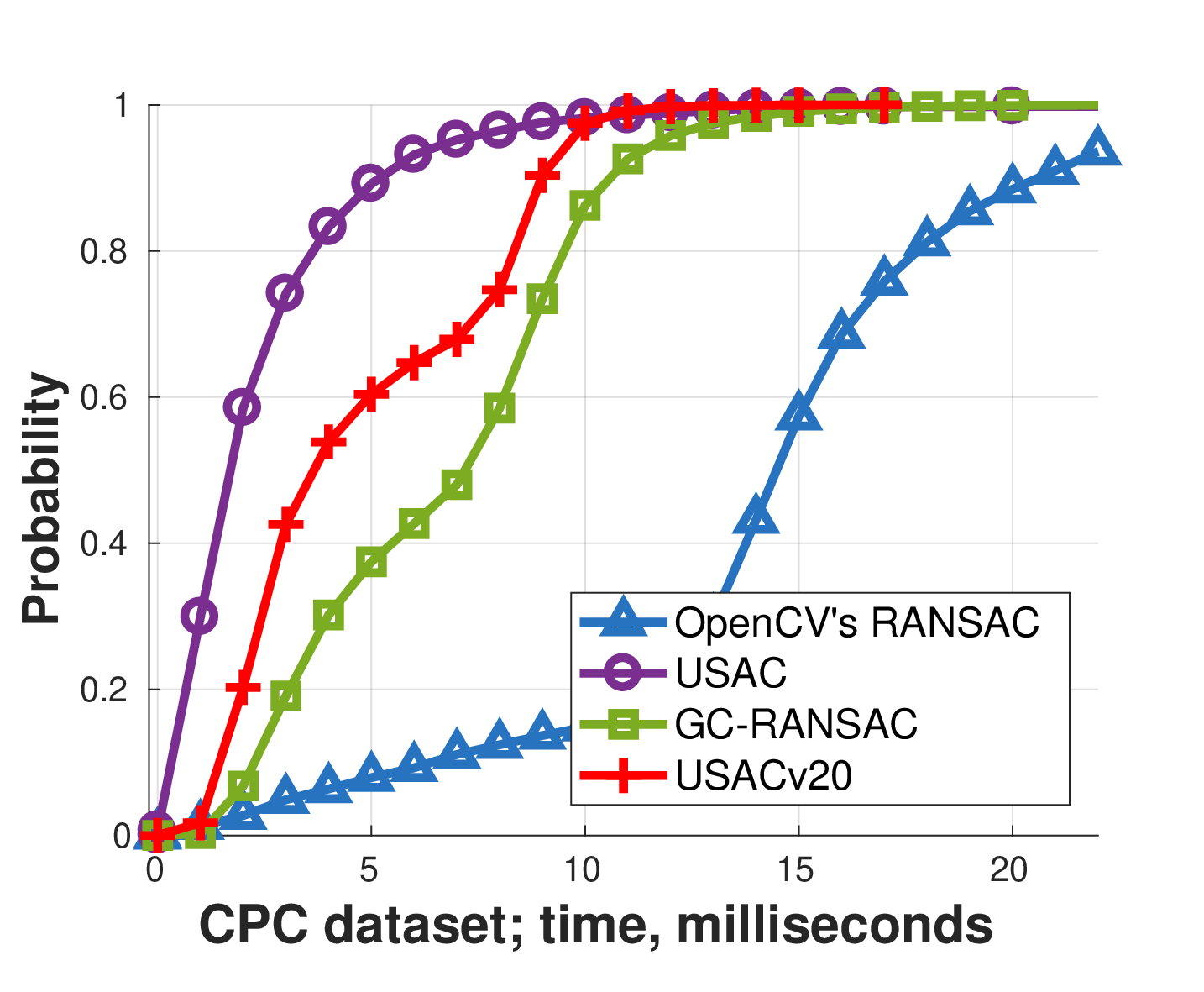}
  \caption{The cumulative distribution functions (CDF) of the Sampson errors (left plot; horizontal axis) and processing times (right; milliseconds) of the estimated fundamental matrices on the \dataset{CPC} dataset.}
  \label{fig:F_cpc}
\end{figure}

\begin{figure}[t]
  \centering
  \includegraphics[width=0.49\columnwidth]{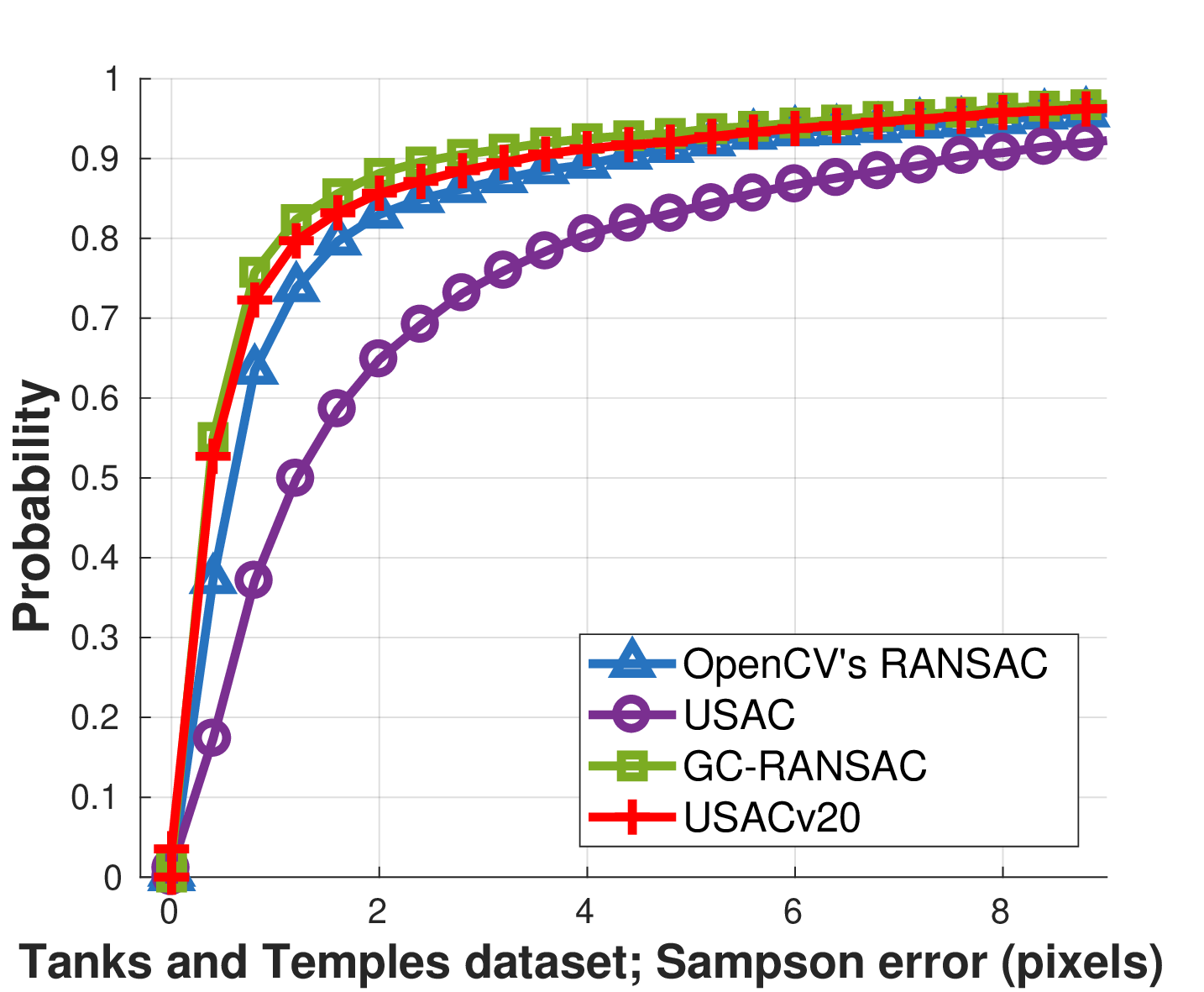}
  \includegraphics[width=0.49\columnwidth]{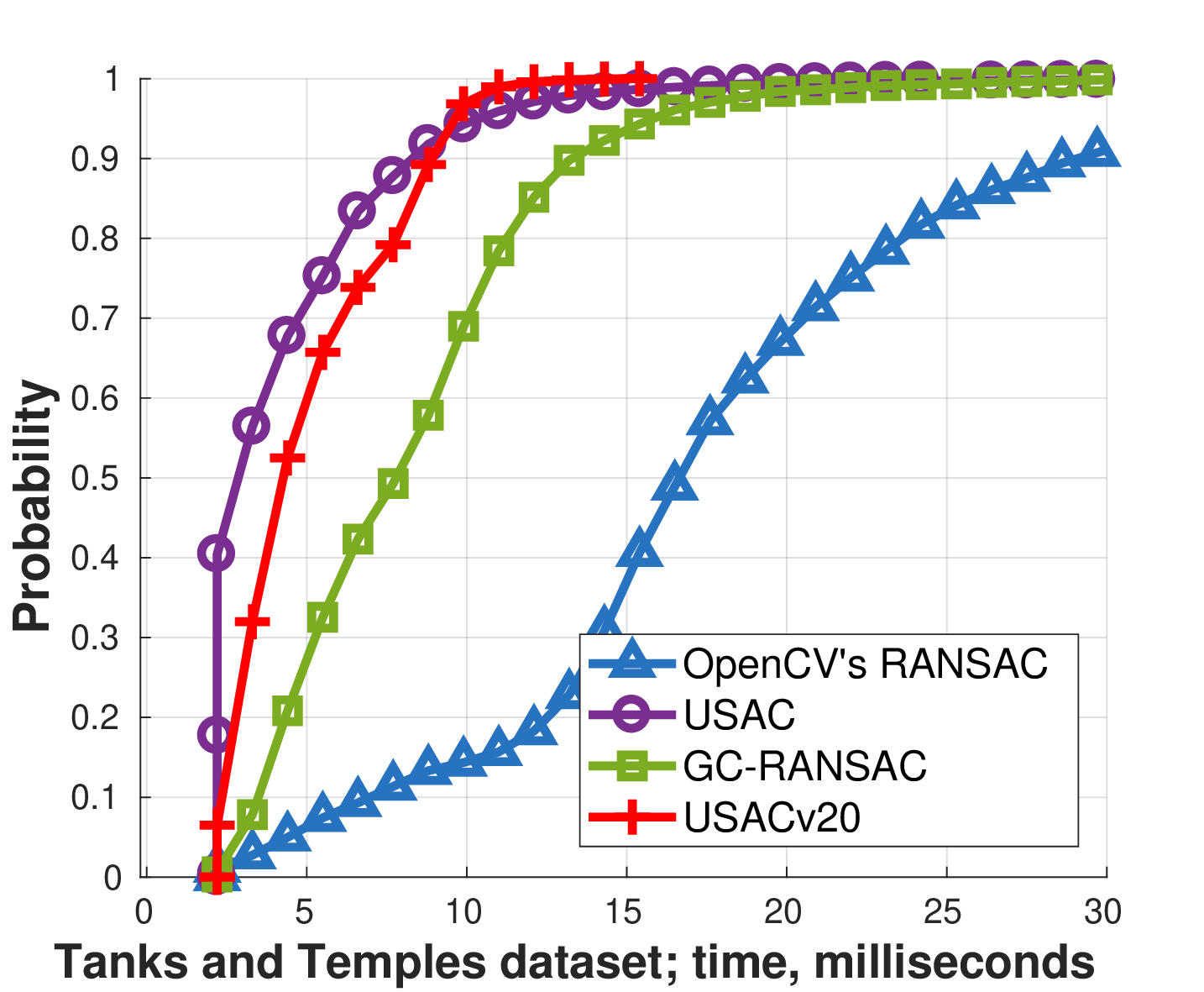}
  \caption{The cumulative distribution functions (CDF) of the Sampson errors (left plot; horizontal axis) and processing times (right; milliseconds) of the estimated fundamental matrices on the \dataset{Tanks and temples} dataset.}
  \label{fig:F_tt}
\end{figure}

In Figures~\ref{fig:F_kitti},\ref{fig:F_tum},\ref{fig:F_tt},\ref{fig:F_cpc}, the cumulative distribution functions (CDF) of the  Sampson  errors  (left plot; horizontal axis) and processing times (right; in milliseconds) of the estimated fundamental matrices are shown.
Being accurate or fast is interpreted by a curve close to the top.
It can be seen that USACv20 is always amongst the top performing methods in terms of geometric accuracy.
The only methods which are faster than USACv20 on any dataset, are significantly less accurate on that particular dataset.
For instance, on \dataset{Tanks and Temples} (Fig.~\ref{fig:F_tt}), USACv20 is the second fastest method (right plot) right after USAC which is the least accurate one (left).

\customparagraph{For homography} estimation, we downloaded \dataset{homogr} (12 pairs) and \dataset{EVD} (15 pairs) datasets~\cite{lebeda2012fixing}. 
They consist of image pairs of different sizes from $329 \times 278$ up to $1712 \times 1712$ with point correspondences and inliers selected manually.   
The \dataset{homogr} dataset contains mostly short baseline stereo images, whilst the pairs of \dataset{EVD} undergo an extreme view change, i.e., wide baseline or extreme zoom. 
In both datasets, the correspondences are assigned manually to one of the two classes, i.e., outlier or inlier of the most dominant homography present in the scene. 
All algorithms applied the normalized four-point algorithm~\cite{hartley2003multiple} for homography estimation and were repeated $100$ times on each image pair. 
To measure the quality of the estimated homographies, we used the RMSE re-projection error calculated from the provided ground truth inliers. 

The fifth and sixth blocks of Table~\ref{tab:magsac_real_experiments} report the median errors ($\epsilon_{\text{med}}$, in pixels), the failure rates ($f$; in percentage) and processing times ($t$; in milliseconds) on the datasets used for homography estimation.
We report the median values to avoid being affected by the failures -- which are also shown. 
A test is considered failure if the error of the estimated model is bigger than the $1\%$ of the image diagonal.
The best values are shown in red, the second best ones are in blue. 
It can be seen that USACv20 is the most accurate method on the \dataset{Homogr} dataset and the second most accurate one on \dataset{ExtremeView}. Its failure ratio and processing time are always the lowest or the second lowest. 

\begin{figure}[t]
  \centering
  \includegraphics[width=0.49\columnwidth]{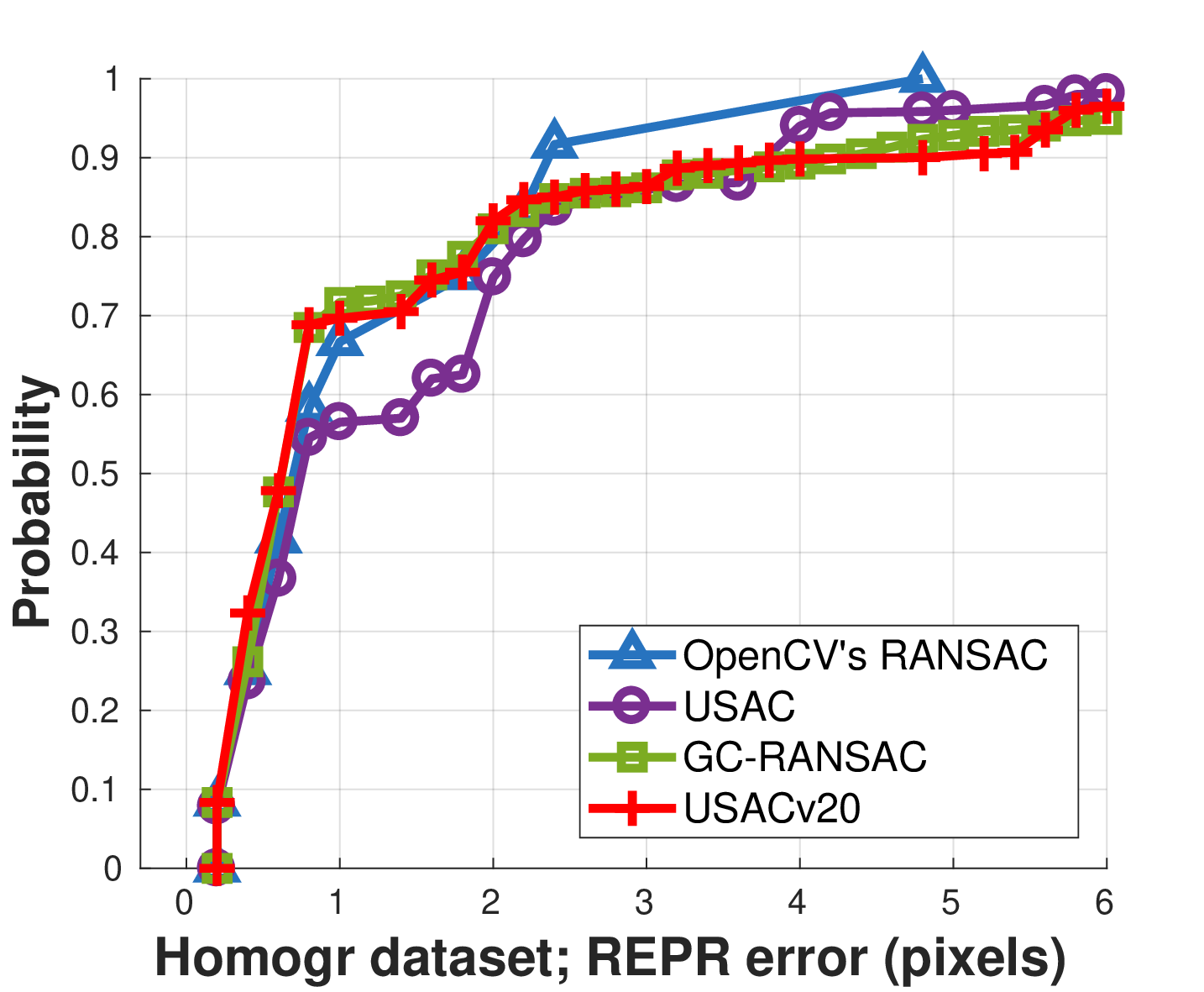}
  \includegraphics[width=0.49\columnwidth]{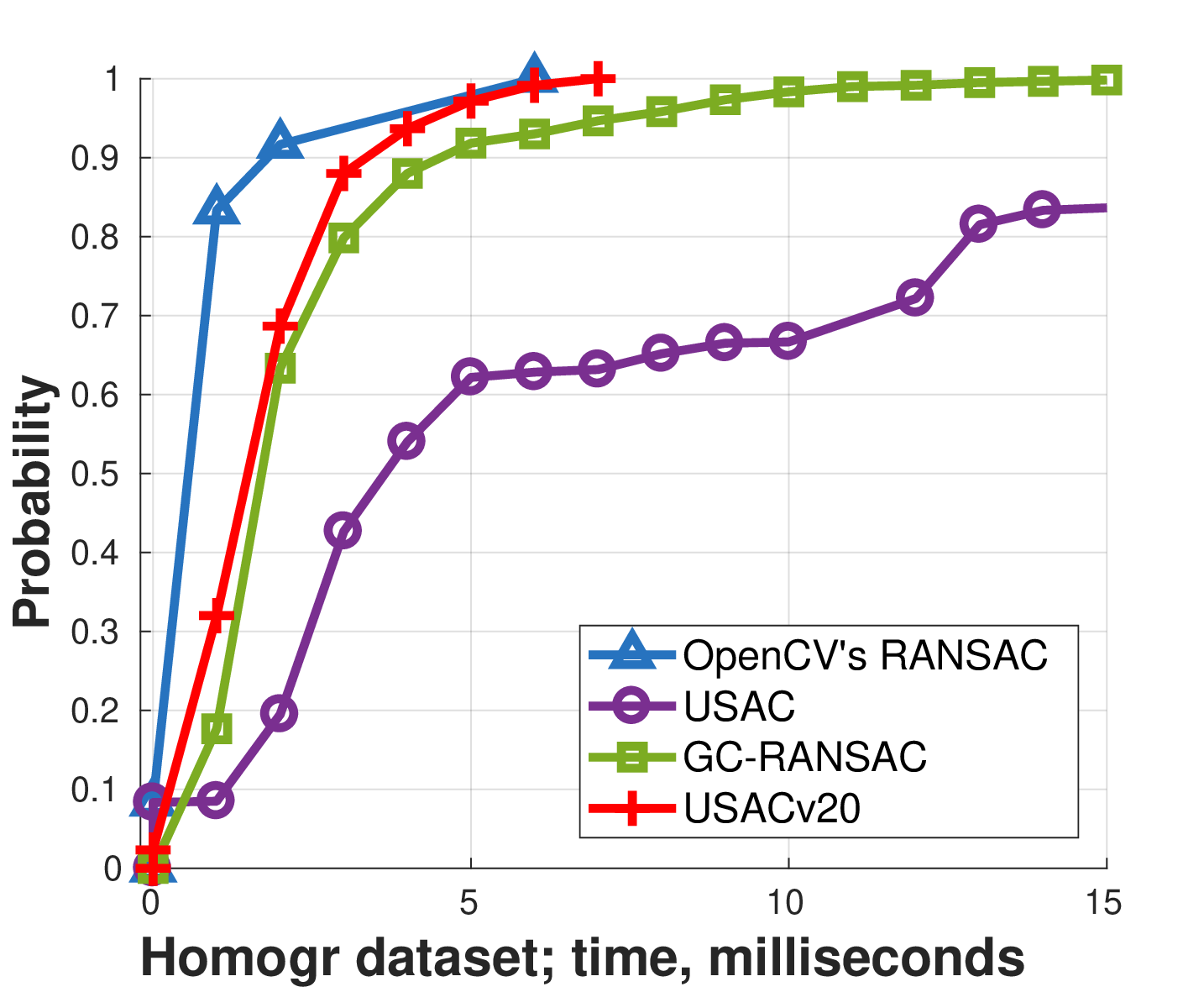}
  \caption{The cumulative distribution functions (CDF) of the Re-projection errors (left plot; horizontal axis) and processing times (right; milliseconds) of the estimated homographies on the \dataset{Homogr} dataset.}
  \label{fig:H_homogr}
\end{figure}

\begin{figure}[t]
  \centering
  \includegraphics[width=0.49\columnwidth]{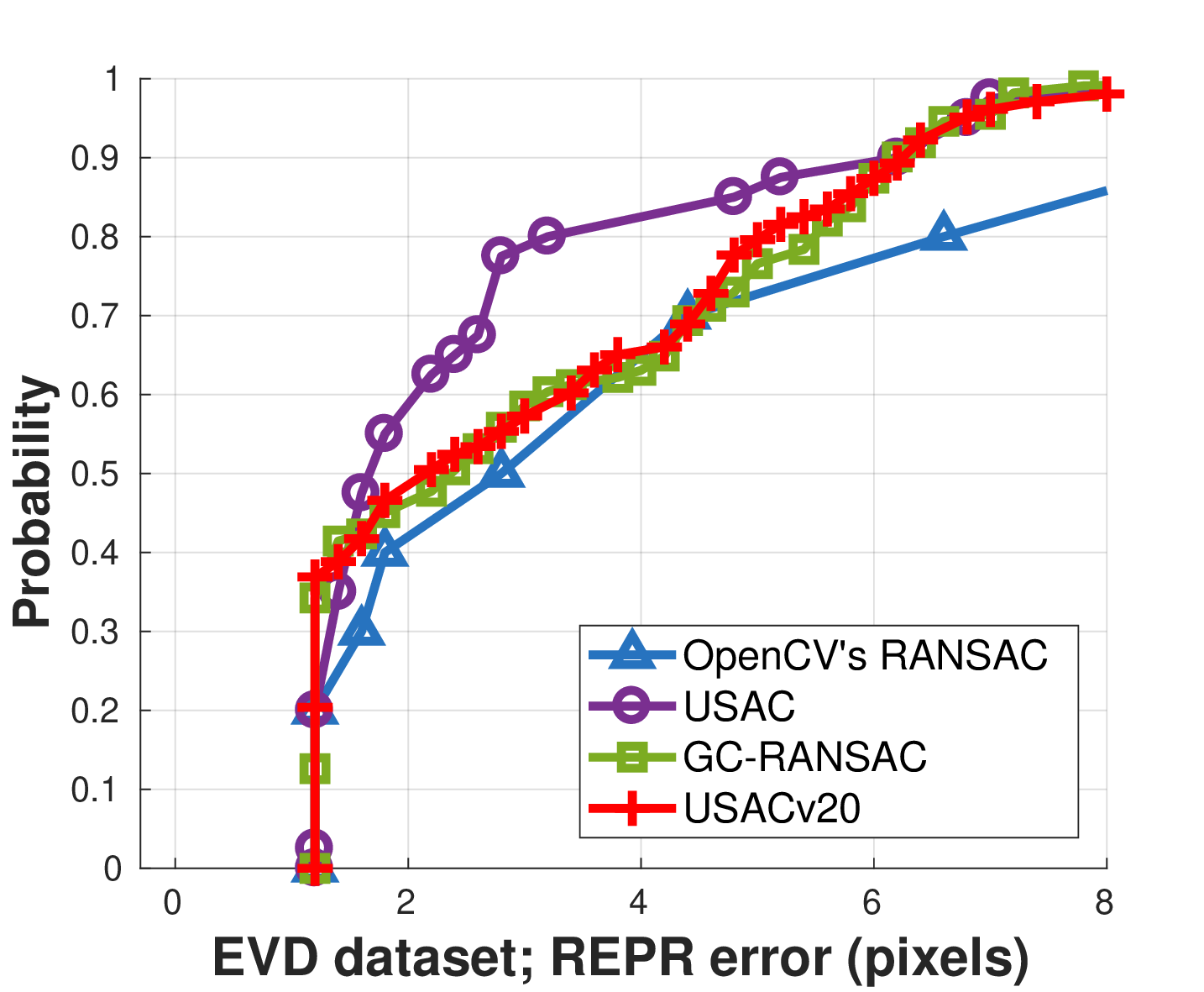}
  \includegraphics[width=0.49\columnwidth]{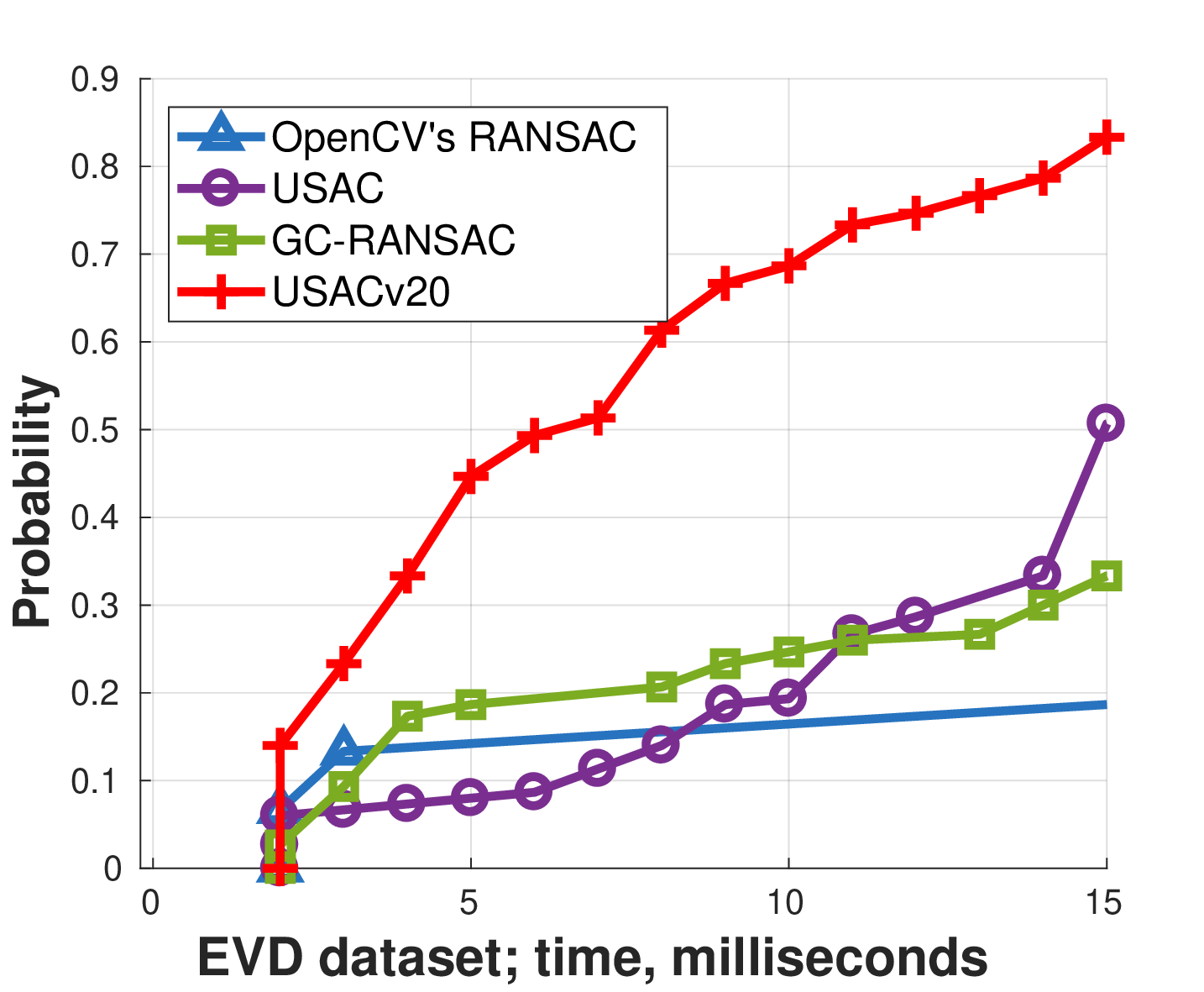}
  \caption{The cumulative distribution functions (CDF) of the Re-projection errors (left plot; horizontal axis) and processing times (right; milliseconds) of the estimated homographies on the \dataset{ExtremeView} dataset.}
  \label{fig:H_evd}
\end{figure}

In Figures~\ref{fig:H_homogr},\ref{fig:H_evd}, the cumulative distribution functions (CDF) of the re-projection errors (left plot; horizontal axis) and processing times (right; in milliseconds) of the estimated homographies are shown.
Being accurate or fast is interpreted by a curve close to the top.
It can be seen that USACv20 is always amongst the most accurate methods.
Its processing time is the second best on \dataset{Homogr} dataset by a margin of 2-3 ms. On \dataset{ExtremeView}, USACv20 is significantly faster than all the competitor robust estimators.

\customparagraph{For essential matrix} estimation, we downloaded the \dataset{Strecha} (1359 pairs) dataset and the \dataset{Piccadilly} scene from the \dataset{1DSfM} dataset\footnote{\url{http://www.cs.cornell.edu/projects/1dsfm/}} \cite{wilson2014robust}. 
For the images of \dataset{Strecha}, both the intrinsic camera parameters and the ground truth poses are provided. 
First, we detected SIFT correspondences~\cite{lowe1999object}, filtered them by the standard SNN ratio test~\cite{lowe1999object}
The intrinsic parameters were used for normalizing the point coordinates.
The ground truth pose was used for validation purposes selecting the ground truth inlier correspondences from the detected ones. These selected inliers were then used for measuring the error of the estimated essential matrices. 
The \dataset{1DSfM} dataset consists of 13 scenes of landmarks with photos of varying sizes collected from the internet. 
It provides 2-view matches with epipolar geometries and a reference reconstruction from incremental SfM (computed with Bundler~\cite{snavely2006photo,snavely2008modeling}) for measuring error. 
We iterated through the provided 2-view matches, detected SIFT correspondences~\cite{lowe1999object}, filtered them by the standard SNN ratio test~\cite{lowe1999object}, and calculated the ground truth relative pose from the reference reconstruction made by the Bundler algorithm.
Note that all image pairs were excluded from the evaluation where fewer than $20$ correspondences were found. 
For the evaluation, we chose the largest scene, i.e.\ Piccadilly, consisting of $7,351$ images.

The last two blocks of Table~\ref{tab:magsac_real_experiments} report the median errors ($\epsilon_{\text{med}}$, in pixels), the failure rates ($f$; in percentage) and processing times ($t$; in milliseconds) on the datasets used for essential matrix estimation.
The best values are shown in red, the second best ones are in blue. 
It can be seen that USACv20 is the most accurate method on both datasets while being the second fastest one. 

\begin{figure}[t]
  \centering
  \includegraphics[width=0.49\columnwidth]{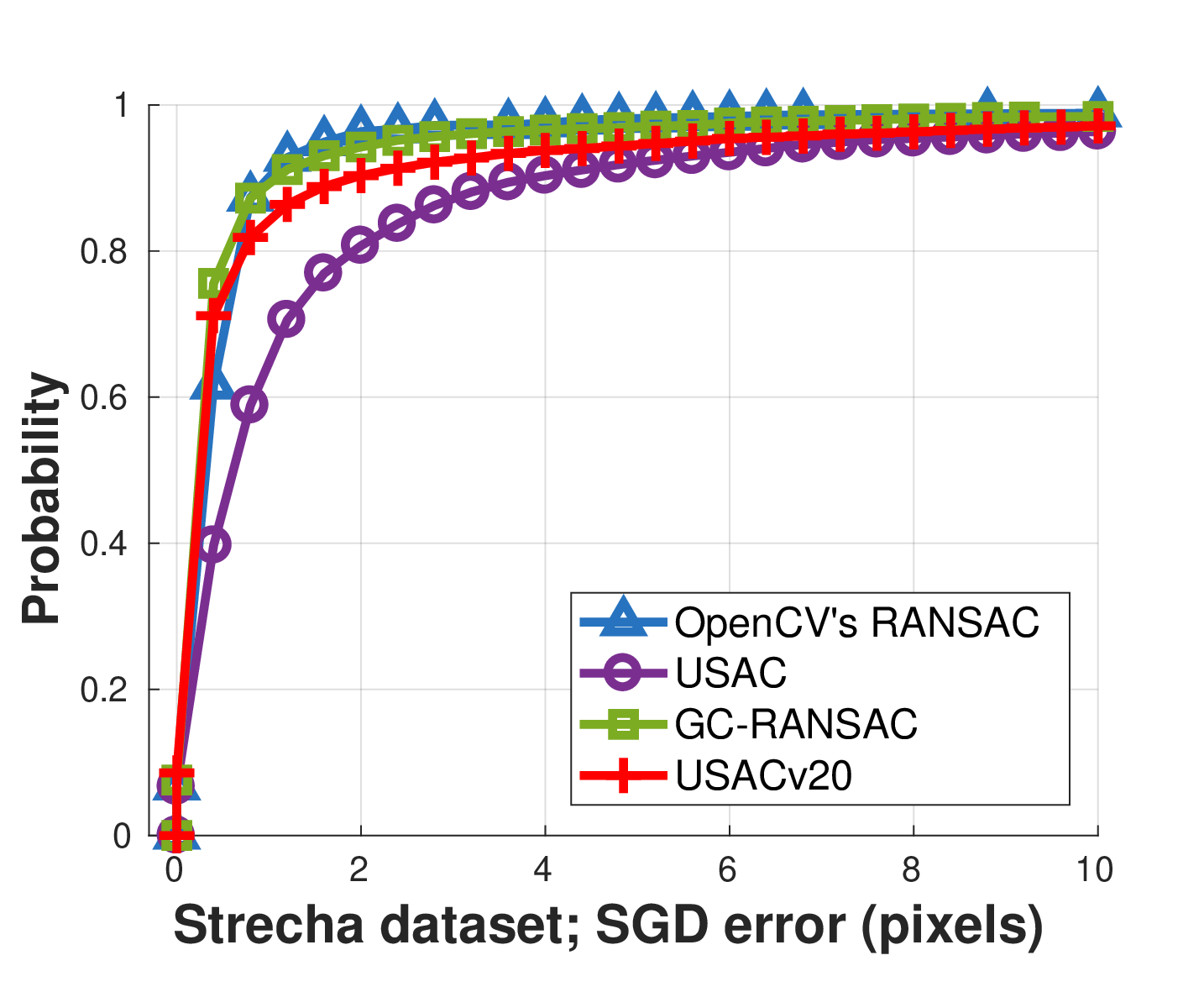}
  \includegraphics[width=0.49\columnwidth]{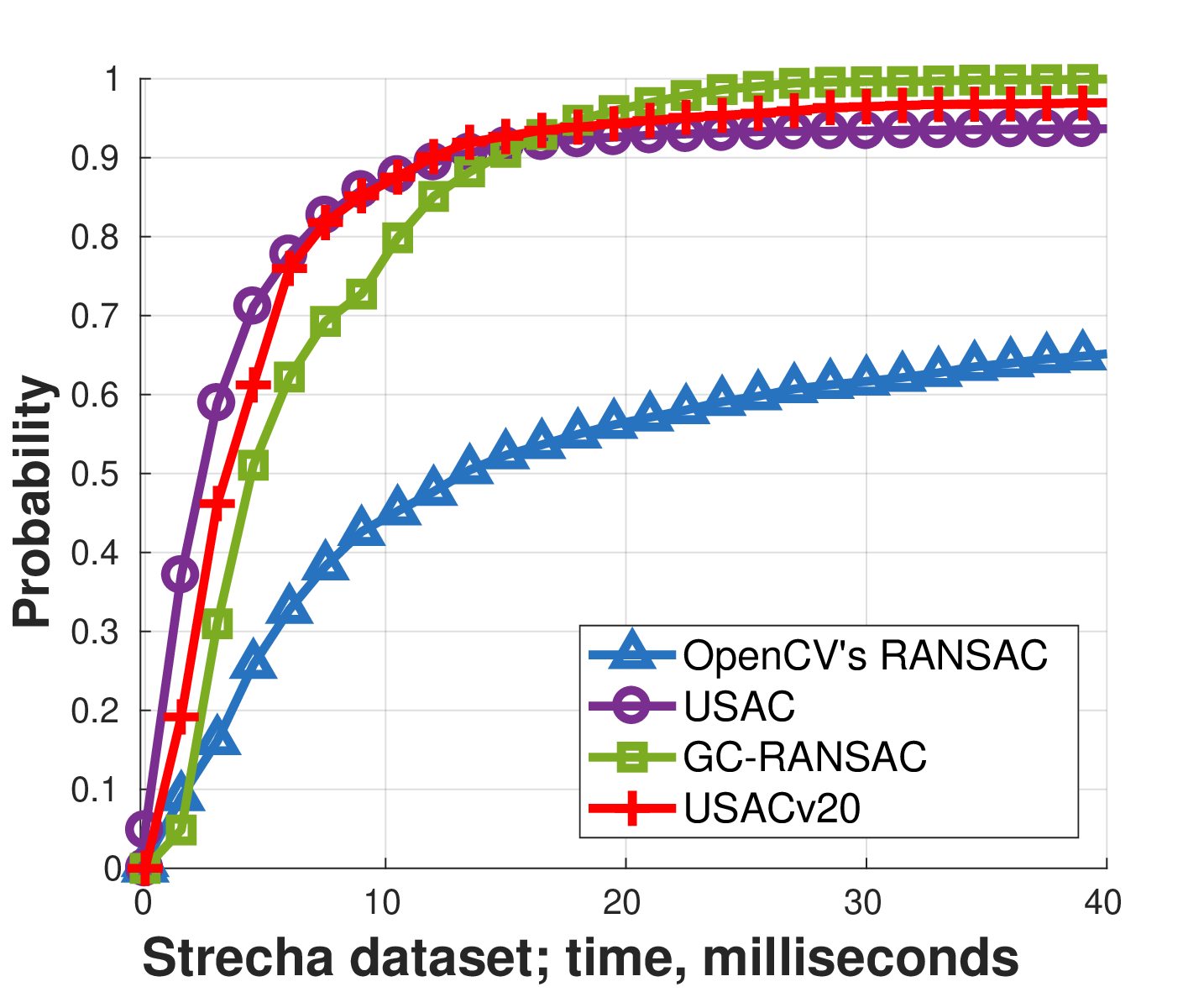}
  \caption{The cumulative distribution functions (CDF) of the Sampson errors (left plot; horizontal axis) and processing times (right; milliseconds) of the estimated essential matrices on the \dataset{Strecha} dataset. }
  \label{fig:E_strecha}
\end{figure}

\begin{figure}[t]
  \centering
  \includegraphics[width=0.49\columnwidth]{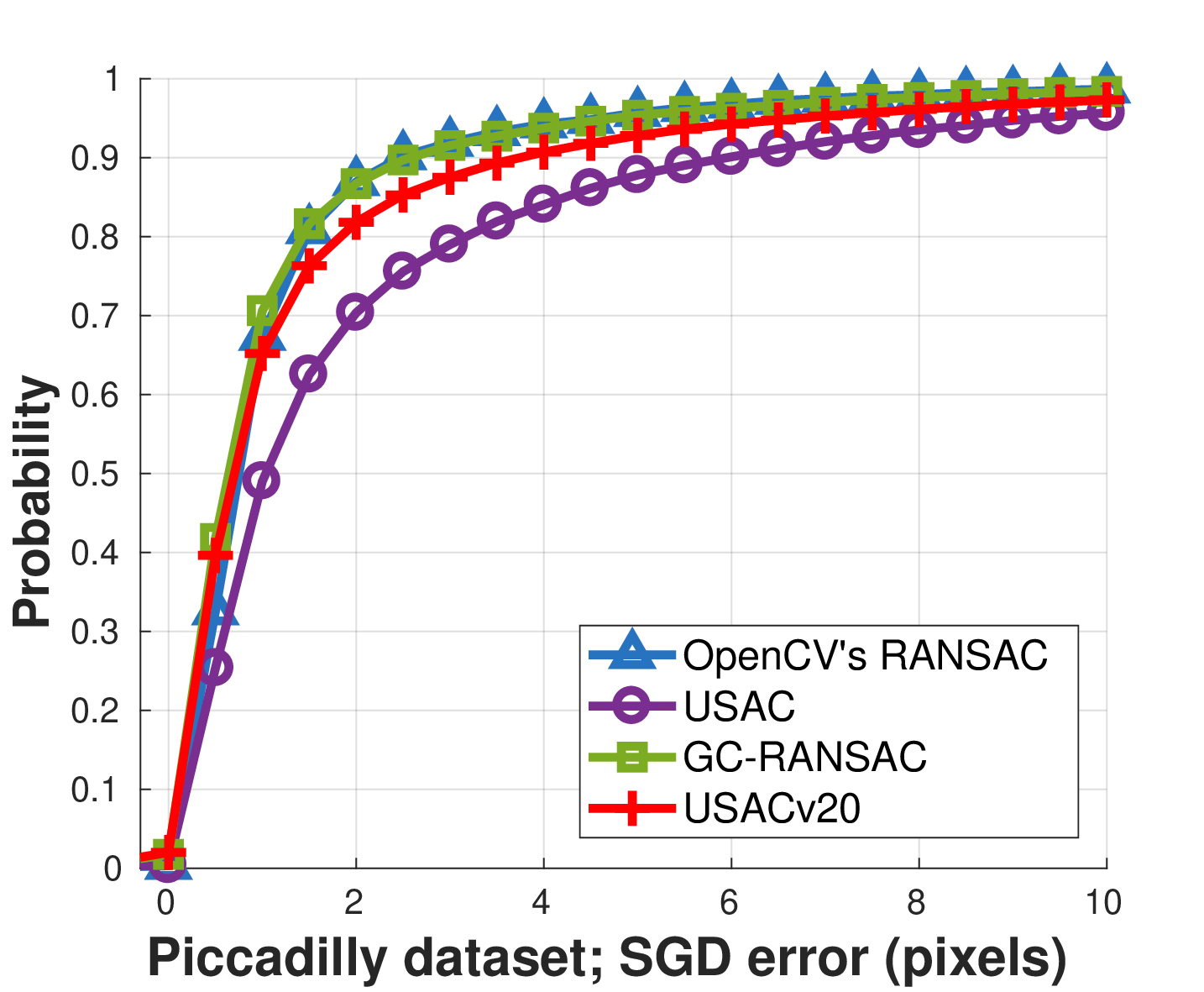}
  \includegraphics[width=0.49\columnwidth]{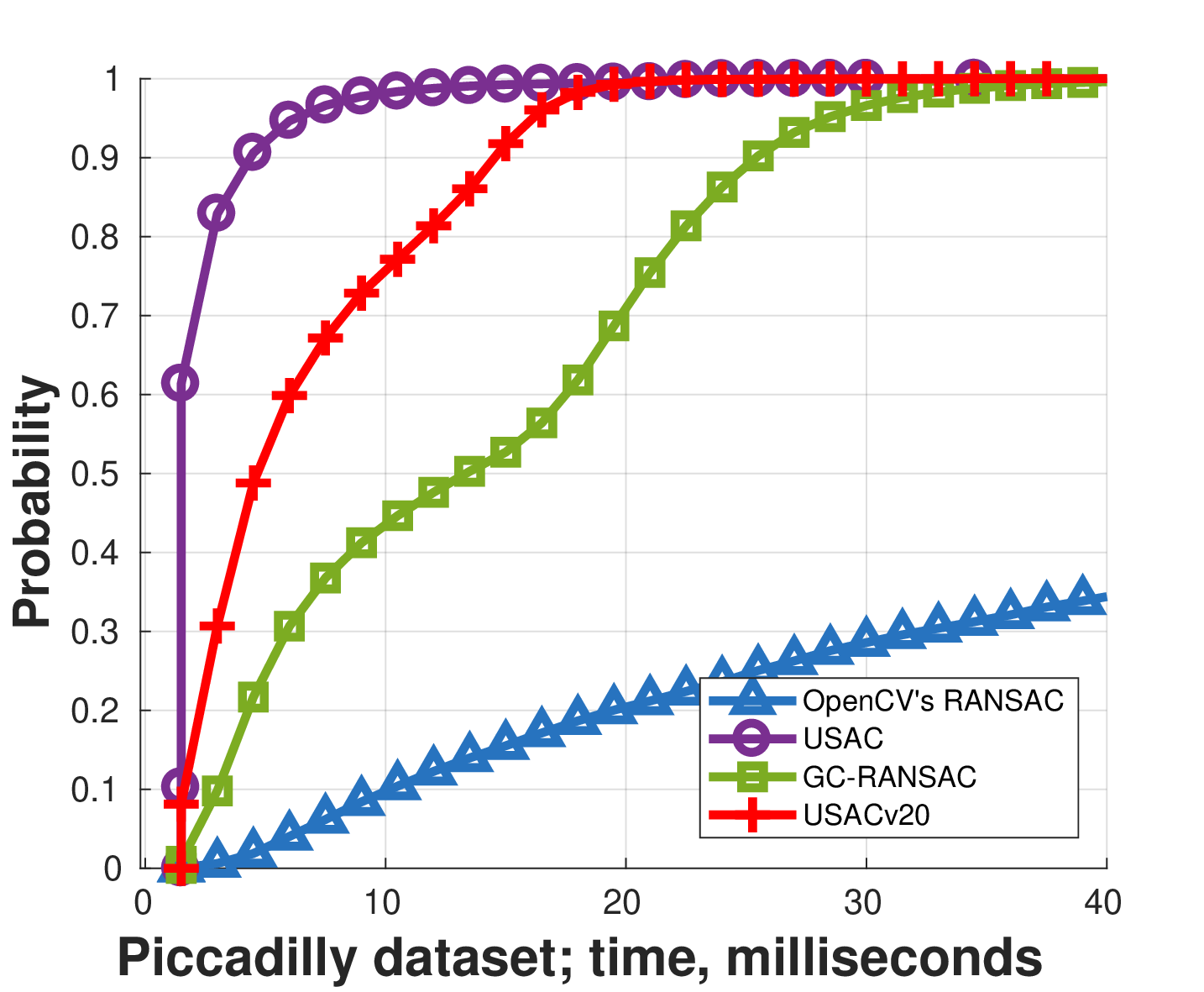}
  \caption{The cumulative distribution functions (CDF) of the Sampson errors (left plot; horizontal axis) and processing times (right; milliseconds) of the estimated essential matrices on the Piccadilly scene of the \dataset{1DSfM} dataset. }
  \label{fig:E_picc}
\end{figure}

In Figures~\ref{fig:E_strecha},\ref{fig:E_picc}, the cumulative distribution functions (CDF) of the SGD errors (left plot; horizontal axis) and processing times (right; in milliseconds) of the estimated homographies are shown.
Being accurate or fast is interpreted by a curve close to the top.
It can be seen that USACv20 is always amongst the most accurate methods while being marginally slower than USAC. However, since USAC does not have essential matrix solver so only fundamental matrices were estimated and then converted to essential matrix using ground truth intrinsic matrices. In general, 5-points algorithm \cite{nister2004efficient} is much slower than 7-points algorithm which was used for $F$-estimation and number of output models for $E$ ranges from 0 to 10 while number of estimated $F$ matrices is at most 3; consequently all of these makes USAC framework faster.

\customparagraph{In summary}, the proposed USACv20 is, on all but one dataset (i.e., \dataset{ExtremeView}), more accurate than the original USAC algorithm while, usually, being faster. Even though USAC is more accurate on \dataset{ExtremeView}, it fails twice as often as USACv20.

The values reported in Table~\ref{tab:magsac_real_experiments} are summarized in Table~\ref{tab:average_over_all}.
It can be seen that the proposed algorithm is, on average, more accurate and faster than the compared state-of-the-art robust estimators. 
Its failure rate is the second best right behind GC-RANSAC. 

\begin{table*}[t!]
   \centering
   \resizebox{0.9999\textwidth}{!}{\begin{tabular}{ l c c c | c c c | c c c | c c c | c c c | c c c | c c c | c c c }
      \Xhline{2\arrayrulewidth}
      & \multicolumn{12}{c}{Fundamental matrix} & \multicolumn{6}{c}{Homography} & \multicolumn{6}{c}{Essential matrix} \\
      \cdashlinelr{1-25}
      & \multicolumn{3}{c}{\dataset{KITTI}~\cite{geiger2012we}} & \multicolumn{3}{c}{\dataset{TUM}~\cite{sturm2012benchmark}} & \multicolumn{3}{c}{\dataset{T\&T}~\cite{knapitsch2017tanks}} & \multicolumn{3}{c}{\dataset{CPC}~\cite{wilson2014robust}} & \multicolumn{3}{c}{\dataset{Homogr}~\cite{lebeda2012fixing}} & \multicolumn{3}{c}{\dataset{EVD}~\cite{lebeda2012fixing}} & 
      \multicolumn{3}{c}{\dataset{Strecha~\cite{strecha2004wide}}} &
      \multicolumn{3}{c}{\dataset{Piccadily~\cite{wilson2014robust}}} \\
      \hline
      \multicolumn{1}{l |}{} & $\epsilon_{\text{med}}$ & $t$ & $f(\%)$  & $\epsilon_{\text{med}}$ & $t$ & $f$ & $\epsilon_{\text{med}}$ & $t$ & $f$ & $\epsilon_{\text{med}}$ & $t$ & $f$ & $\epsilon_{\text{med}}$ & $t$ & $f$ & $\epsilon_{\text{med}}$ &
       $t$ & $f$ & $\epsilon_{\text{med}}$ &
       $t$ & $f$ & $\epsilon_{\text{med}}$ &
       $t$ & $f$ \\
      \hline 

      \multicolumn{1}{l |}{\Rowcolor{lightgray} USACv20} & \win{0.2} & \second{1.9} & \second{0.2} & \win{0.3} & \win{2.1} & \second{8.4} & \win{0.6} & \second{5.6} & \win{12.9} & \win{0.5} & \second{5.3} & \second{43.0} & \win{0.7} & \second{2.2} & \win{0.0} & \second{2.3} & \win{8.5} & \second{31.3} & \win{0.4} & \second{8.1} & 4.6 & \win{0.9} & \second{7.3} & \second{2.2} \\
      
      \multicolumn{1}{l |}{\Rowcolor{lightgray} GC-RANSAC} & \second{0.3} & 2.3 & \win{0.1} & \second{0.4} & 3.1 & 8.6 & \win{0.6} & 8.8 & \second{13.0} & \win{0.5} & 7.2 & \win{42.8} & \second{0.8} & 2.8 & \win{0.0} & 2.5 & \second{24.5} & \win{26.0} & \win{0.4} & \win{7.4} & \second{3.8} & \win{0.9} & 14.5 & 3.1  \\
      
      \multicolumn{1}{l |}{\Rowcolor{lightgray} USAC} & 0.4 & 3.3 & 0.3 & 0.6 & \second{2.2} & 9.2 & 1.4 & \win{4.4} & 15.0 & 0.8 & \win{3.1} & 44.0 & 0.9 & 10.0 & \win{0.0} & \win{1.8} & 25.0 & 73.3 & 0.8 & 9.1 & \second{3.8} & 1.3 & \win{2.6} & 3.1 \\
      
      \multicolumn{1}{l |}{\Rowcolor{lightgray} OpenCV} & 0.4 & \win{1.6} & \second{0.2} & 0.5 & 4.4 & \win{8.3} & 0.8 & 18.5 & \second{13.0} & 0.7 & 14.9 & 45.2 & 0.9 & \win{1.3} & \win{0.0} & 3.5 & 136.0 & 33.3 & 0.5 & 69.2 & \win{3.0} & 1.0 & 121.0 & \win{0.8}   \\
      \Xhline{2\arrayrulewidth}
   \end{tabular}}
   \caption{Median errors ($\epsilon_{\text{med}}$), failure rates ($f$; as percentages) and avg.\ run-times ($t$, in milliseconds) are reported for each method on all tested problems and datasets. 
   The error of the fundamental matrices is the Sampson distance from the ground truth. 
   For homographies, the RMSE re-projection error from ground truth inliers is used.
   For essential matrix, the error is symmetric geometric distance (SGD) of normalized points.
   A test is considered a  failure if the error is bigger than $1\%$ of the image diagonal.
   For each method, the inlier-outlier threshold was set to maximize the accuracy (for fundamental matrix is 1 pixel, for homographies 2 pixels and for essential matrix, 1 pixel normalized by the intrinsic matrices) and the confidence to $0.99$. 
   The best values in each column are shown by red and the second best ones by blue.}
    \label{tab:magsac_real_experiments}
\end{table*}

\begin{table}[t!]
   \centering
   \resizebox{0.8\columnwidth}{!}{\begin{tabular}{ l | c c c c }
      \Xhline{2\arrayrulewidth}
      & USACv20 & GC-RANSAC & USAC & OpenCV \\
      \hline 
      $\epsilon$ & \win{0.7} & \second{0.8} & 1.0 & 1.0 \\
      $t$ & \win{5.1} & 8.8 & \second{7.5} & 45.9 \\
      $f$ & \second{12.8} & \win{11.9} & 18.6 & 13.0 \\
      \Xhline{2\arrayrulewidth}
   \end{tabular}}
   \caption{The avg.\ of the errors ($\epsilon$; in pixels), processing times ($t$; in milliseconds) and failure rates ($f$; in percentages) in Table~\ref{tab:magsac_real_experiments} are reported. The best values in each column are shown by red and the second best ones by blue. }
    \label{tab:average_over_all}
\end{table}

\section{Conclusion}

In this paper, we reviewed some of the most recent RANSAC variants, combined them together and proposed a state-of-the-art variant, i.e.\ USACv20, of the Universal Sample Consensus~\cite{raguram2013usac} (USAC) algorithm. 
USACv20 is tested on 8 datasets, estimating homographies, fundamental and essential matrices. 
On average, it leads to the most geometrically accurate models and it is fastest compared to USAC, OpenCV's RANSAC and Graph-Cut RANSAC. 
Compared to the original USAC, all reported properties improved significantly. 
Also, an important objective was to implement a modular and optimized framework in C++ to make future RANSAC modules easy to be combined with.
The pipeline will be made available after publication.

\section{Acknowledgement}
This research was supported by Czech Technical University student grant SGS OHK3-019/20.

{\small
\bibliographystyle{ieee.bst}
\bibliography{egbib.bib}
}

\end{document}